\pdfoutput=1

\documentclass[11pt]{article}

\usepackage[final]{acl}

\usepackage{times}
\usepackage{latexsym}

\usepackage[T1]{fontenc}

\usepackage[utf8]{inputenc}

\usepackage{microtype}

\usepackage{inconsolata}

\usepackage{graphicx}

\usepackage{footmisc}
\usepackage{float}
\usepackage{enumitem}
\usepackage{multirow}
\usepackage{makecell}
\usepackage{colortbl}
\usepackage[most]{tcolorbox}
\usepackage{booktabs}
\usepackage{arydshln} 
\usepackage{wrapfig}
\usepackage{amssymb}
\usepackage{amsmath}
\usepackage{soul}
\usepackage{xcolor}
\definecolor{myblue0}{RGB}{192, 214, 234}
\definecolor{myblue1}{RGB}{166, 199, 226}
\definecolor{mygray}{RGB}{224, 224, 232}
\newcommand{\mysize}{\fontsize{9.8pt}{9.8pt}\selectfont}
\newcommand{\mysizeone}{\fontsize{10.2pt}{12pt}\selectfont}

\usepackage{stfloats}

\tcbuselibrary{breakable}
\usepackage[linesnumbered,ruled,vlined,algo2e]{algorithm2e}
\usepackage{algorithmic}
\usepackage{algorithm}
\newcommand{\mycommentstyle}[1]{\color[HTML]{0671b9}{\small #1}}
\SetKwComment{Comment}{\mycommentstyle{// }}{}
\SetKwComment{inComment}{\mycommentstyle{}}{}

\usepackage{etoolbox}
\makeatletter
\renewcommand\@fnsymbol[1]{\ensuremath{\ifcase#1\or \dagger\or *\or \ddagger\or \mathsection\or \mathparagraph\or \|\or **\or \dagger\dagger\or \ddagger\ddagger \else\@ctrerr\fi}}
\makeatother

\title{PGPO: Enhancing Agent Reasoning via Pseudocode-style \\Planning Guided Preference Optimization}

\author{
 \textbf{Zouying Cao}$^{1,3,4,}$\thanks{Work done during an internship at Alibaba Group.},
 \textbf{Runze Wang}$^{2}$,
 \textbf{Yifei Yang}$^{1,3,4}$,
 \textbf{Xinbei Ma}$^{1,3,4}$,
 \\
 \textbf{Xiaoyong Zhu}$^{2}$,
 \textbf{Bo Zheng}$^{2,}$\thanks{Corresponding authors. This research was supported by the Joint Research Project of Yangtze River Delta Science and Technology Innovation Community (No. 2022CSJGG1400), Alibaba Group through Alibaba Innovative Research Program.
 },
 \textbf{Hai Zhao}$^{1,3,4,}$\footnotemark[2]
\\
 \textsuperscript{1}School of Computer Science, Shanghai Jiao Tong University, \\
 \textsuperscript{2}Taobao \& Tmail Group of Alibaba, 
 \textsuperscript{3}Key Laboratory of Shanghai Education Commission \\for Intelligent Interaction and Cognitive Engineering, Shanghai Jiao Tong University,\\
 \textsuperscript{4}Shanghai Key Laboratory of Trusted Data Circulation and Governance in Web3
\\
\texttt{zouyingcao@sjtu.edu.cn, yunze.wrz@alibaba-inc.com}
}

\begin{document}
\maketitle
\begin{abstract}
Large Language Model (LLM) agents have demonstrated impressive capabilities in handling complex interactive problems. 
Existing LLM agents mainly generate natural language plans to guide reasoning, which is verbose and inefficient. 
NL plans are also tailored to specific tasks and restrict agents' ability to generalize across similar tasks.
To this end, we explore pseudocode-style plans (P-code Plan) to capture the structural logic of reasoning. 
We find that P-code Plan empowers LLM agents with stronger generalization ability and more efficiency.  
Inspired by this finding, we propose a pseudocode-style \underline{P}lanning \underline{G}uided \underline{P}reference \underline{O}ptimization method called PGPO for effective agent learning.
With two planning-oriented rewards, PGPO further enhances LLM agents' ability to generate high-quality P-code Plans and subsequent reasoning.
Experiments show that PGPO achieves superior performance on representative agent benchmarks and outperforms the current leading baselines. 
Analyses reveal the advantage of PGPO in reducing action errors and omissions during reasoning.\footnote{\url{https://github.com/zouyingcao/PGPO}.} 
\end{abstract}

\section{Introduction}
Recent advances in large language models have promoted the development of LLM agents~\cite{wang2024survey,xi2025rise}. 
Planning serves as a critical component of agent reasoning~\cite{huang2024understanding}, allowing agents to break down complex problems into manageable subtasks. 
By prompting strategies~\cite{wang2023plan,prasad2024adapt,roy2024flap}, task-specific fine-tuning~\cite{qiao2024agent,qiao-etal-2024-autoact} or external classical planners~\cite{liu2023llm+,arora2024anticipate}, LLM agents are equipped with basic planning abilities. 
Due to the dominance of natural language (NL) in agent reasoning~\cite{cot,yao2023react}, existing researches mainly focus on generating NL plans. 
However, semantic ambiguities and undesired verbosity in NL may not be beneficial to agent planning, leading to low precision and inefficiency. 
Meanwhile, NL plans are too specific to help LLM agents generalize to other unseen yet similar tasks.
Therefore, it still remains underexplored whether alternative plan formats could elicit more efficient and generalized LLM agents.

\begin{figure}
    \includegraphics[width=1\linewidth]{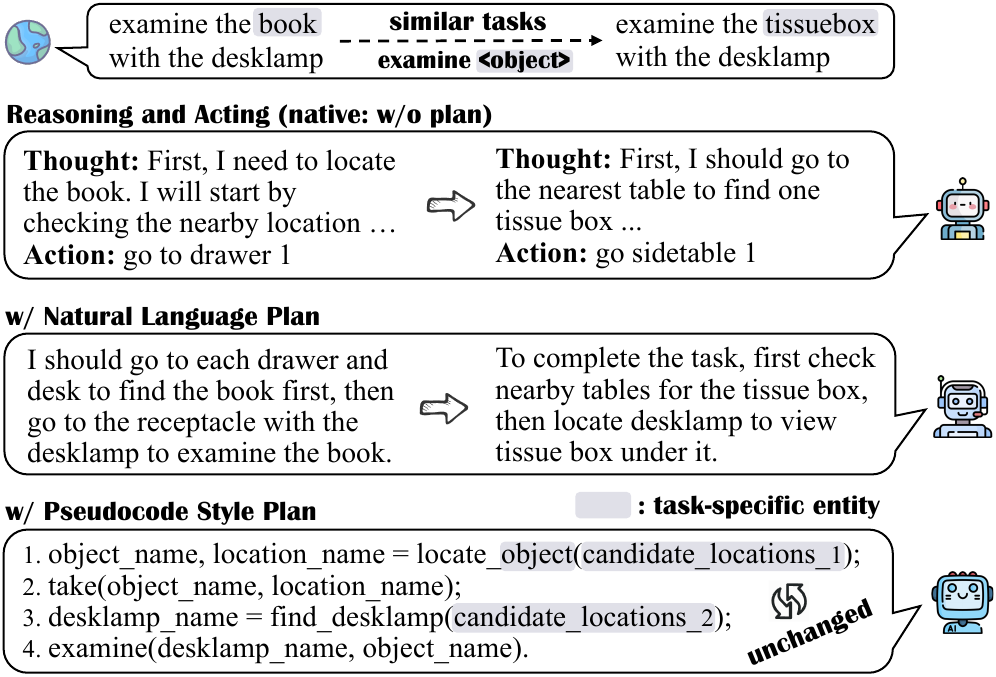}
    \caption {An example demonstrating why P-code Plan helps LLM agents generalize well. When faced with similar tasks (e.g., examine \sethlcolor{mygray}\hl{object} with desklamp), the thought process can be recycled through P-code Plans.}
    \label{fig:p-code_plan}
\end{figure}

\begin{figure*}
    \includegraphics[width=1\linewidth]{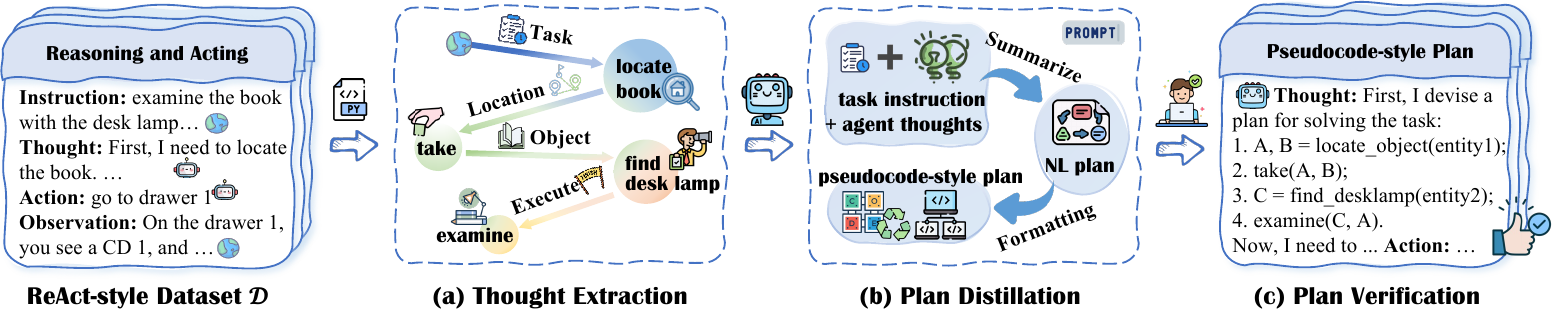}
    \caption{Overview of P-code Plan generation pipeline. We first extract the thought part from existing ReAct-style datasets. Then, we prompt GPT-4o to summarize the thought process into high-level plans. Pseudocode-style plans are finally structured with predefined formats, followed by manual verification to ensure accuracy.}
    \label{fig:plan_gen}
\end{figure*}

Prior work~\cite{codeact} demonstrates the advantage of executable code as agent's action over text or JSON format. 
Inspired by this, we explore using pseudocode to represent plan, given that plans can be considered as high-level abstraction of actions while pseudocode outlines code. 
Our work starts by distilling pseudocode-style plans (denoted as P-code Plan) from existing ReAct-style~\cite{yao2023react} datasets, adhering to predefined format requirements.
We observe that through fine-tuning with P-code Plan incorporated, LLM agents exhibit improved out-of-distribution generalization. 
As shown in Figure~\ref{fig:p-code_plan}, abstract planning steps in P-code Plan can capture generalizable task knowledge while NL plans focus on specific knowledge and may suffer from overfitting. 
Moreover, agent's planning ability learned from concise P-code Plan facilitates a more efficient reasoning process with fewer interactions.

Building upon these insights, we further propose a pseudocode-style \underline{P}lanning \underline{G}uided \underline{P}reference \underline{O}ptimization method named PGPO for agent capability enhancement. 
Specifically, we first utilize supervised fine-tuning to build a base agent. 
Then, PGPO contains two iterative phases: (1) the base agent performs exploration on expert trajectories to construct contrastive trajectory datasets based on two designed planning-oriented rewards; (2) direct preference optimization is employed to refine the base agent's pseudocode-style planning ability for task guidance. 
Through experiments with four LLMs, PGPO outperforms various strong baselines by relative 11.6\% performance gain averaged across three representative agent benchmarks.
In summary, our contributions are as follows:
\begin{itemize}[leftmargin=12pt,topsep=2pt,itemsep=-2pt]%
    \item We investigate the effectiveness of pseudocode-style plans in agent reasoning, which are more concise and structured than NL plans. P-code Plan demonstrates its superiority in boosting the generalization ability of LLM agents.
    \item We further introduce PGPO, a preference optimization method that empowers LLM agents with enhanced reasoning capabilities under the guidance of pseudocode-style plans. 
    \item Experimental results reveal that our PGPO achieves state-of-the-art performance on representative agent benchmarks, especially when dealing with more complex agent tasks.
\end{itemize}

\section{Pseudocode-style Plan is Beneficial}
In this section, we first define the structural representation of P-code Plan and then introduce a plan generation pipeline, followed by preliminary experiments to show the advantage of P-code Plan.

\begin{figure*}
    \includegraphics[width=1\linewidth]{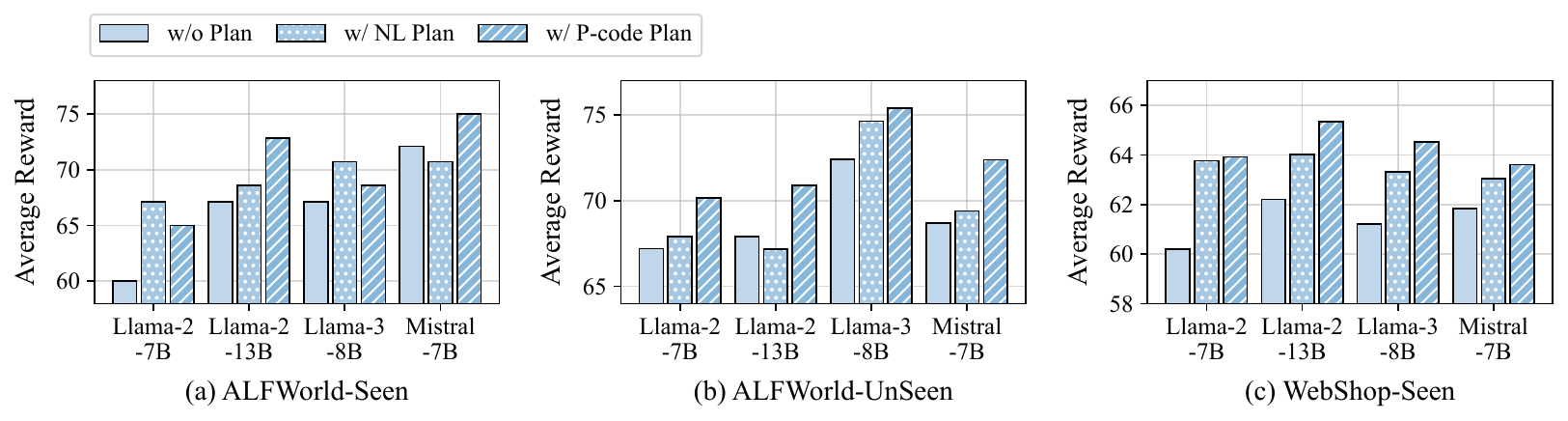}
    \caption {Comparison between \textit{w/ P-code Plan} and \textit{w/o Plan}, \textit{w/ NL Plan} during the SFT process for LLM agents. Here \textit{w/o Plan} symbolizes training on the original ReAct-style datasets, \textit{w/ NL Plan} indicates incorporating natural language plans into training data and \textit{w/ P-code Plan} represents the incorporation of pseudocode-style plans.}
    \label{fig:plan_based_sft}
\end{figure*}

\subsection{Definition of P-code Plan}\label{sec:plan_def}
In this paper, we mainly focus on LLM agents’ multi-step reasoning usage: LLM agents interact with the environment to accomplish complex tasks, which can 
be represented as a set of Thought-Action-Observation  tuples $\{(t, a, o)\}_n$. At each interaction, the LLM agent gives the inner thoughts $t$ and takes an action $a$ based on the observation $o$ from the environment. $n$ denotes the number of interaction turns. 
Following this task formalization, we define the format of our pseudocode-style plans.

\paragraph{Planning Step.} $P_s=(id, name, [parameter], $ $[return\;value], [control\;flow])$ structures each step in the plan, uniquely identified by $id$. 
Similar to function in programming languages, one planning step usually corresponds to a subset of actions oriented to one subtask.
$name$ abstracts a function identifier to describe the reasoning process of this subtask, with $parameters$ enclosed in parentheses.
Square brackets 
$([...])$ mean optional attributes. 
If necessary, $return\;values$ are indicated on the left of the assignment operator, standing for the observed information.
$control\;flow$ signifies standard programming structures such as \textit{if-else} and \textit{for}, which can be omitted when planning steps are performed sequentially.

\paragraph{Planning Entity.} $E=\{e_1, e_2, ..., e_m\}$
refs to the prior knowledge entities used to specify some $parameters$ in the planning step. These entities serve as guidance for generating valid actions and avoiding aimless exploration.

\paragraph{P-code Plan.} Denoted as $(P_s,E)$,  pseudocode-style plans are the combination of abstract planning steps and task-specific planning entities. 
Different from NL plans, P-code Plans are more structured and concise. 
This format helps agent better generalize to unseen tasks\footnote{The term “unseen tasks” refs to task scenarios that are not present in the training set, which can measure out-of-distribution generalization.}, where unseen tasks may share similar planning steps with seen tasks but initialize different planning entities.

\subsection{Plan Generation Pipeline}\label{sec:plan_gen}
Initially, we have ReAct-style dataset $\mathcal{D}$ where each instance $d$ consists of one task instruction $u$ with its collected expert trajectory $\tau=(t_{1}, a_{1}, o_{1},$ $ ..., o_{n-1}, t_{n}, a_{n}, o_{n})$. 
Then, with LLM and human participation, as illustrated in Figure~\ref{fig:plan_gen}, our P-code Plan generation pipeline is outlined as follows:

\begin{itemize}[leftmargin=12pt,topsep=2pt,itemsep=-2pt]
\item \textbf{Thought Extraction.} 
We first extract agent thoughts $\{t\}_n$ from expert trajectories, which involves task-specific knowledge for planning. 
The relationship between agent thoughts and task plans can be viewed as one of abstraction.
The plan distills the essentials of thoughts while omitting specifics, as an abstract summary does. 

\item \textbf{Plan Distillation.} 
Subsequently, to improve the quality of distilled plans and obtain more accurate abstract knowledge, we employ a powerful model (e.g., GPT-4o) for generation. 
Given the task instruction $u$ with agent thoughts $\{t\}_n$, we instruct the LLM to summarize the step-by-step plan in natural language. 
Next, due to the effectiveness of few-shot prompting strategy for structural generation~\cite{valmeekam2024planbench,liu2023llm+}, demonstrations of pseudocode-style plans paired with corresponding tasks are taken as input, converting natural language plans into P-code Plans. 
See Appendix~\ref{app:plan_gen} for the prompt we used to guide plan distillation.

\item \textbf{Plan Verification.} 
Last, we choose human to verify the LLM-generated P-code Plans whether follow the format requirements described in Section~\ref{sec:plan_def}. 
In order to guarantee accuracy and knowledge consistency, minor manual refinement are also needed.

\end{itemize}

\begin{table}
\setlength\tabcolsep{3pt}
  \centering
  \resizebox{\linewidth}{!}{
  \begin{tabular}{lcccc}
   \toprule
   \textbf{Setting}&\textbf{\makecell[c]{Llama-2\\-7B}}&\textbf{\makecell[c]{Llama-2\\-13B}}&\textbf{\makecell[c]{Llama-3\\-8B}}&\textbf{\makecell[c]{Mistral\\-7B}}\\
\midrule
\rowcolor{myblue0!40}\multicolumn{5}{l}{\textbf{ALFWorld-Seen}}\\
   \textit{w/o Plan}&12.04&11.38&11.49&11.45\\
   \textit{w/ NL Plan}&11.34&11.36&11.41&12.14\\
   \textit{w/ P-code Plan}&\textbf{11.33}&\textbf{11.21}&\textbf{11.29}&\textbf{10.99}\\
\midrule
 \rowcolor{myblue0!40}\multicolumn{5}{l}{\textbf{ALFWorld-Unseen}}\\
   \textit{w/o Plan}&12.93&12.22&12.59&12.84\\
   \textit{w/ NL Plan}&12.13&12.27&11.88&12.33\\
   \textit{w/ P-code Plan}&\textbf{12.04}&\textbf{11.87}&\textbf{11.78}&\textbf{11.57}\\
   \bottomrule
  \end{tabular}
  }
  \caption{Average interaction turns required on ALFWorld. \textbf{Bold} indicates the best results of each model.}
  \label{tab:avg_turn}
\end{table}

\begin{figure*}
    \centering
    \includegraphics[width=1\linewidth]{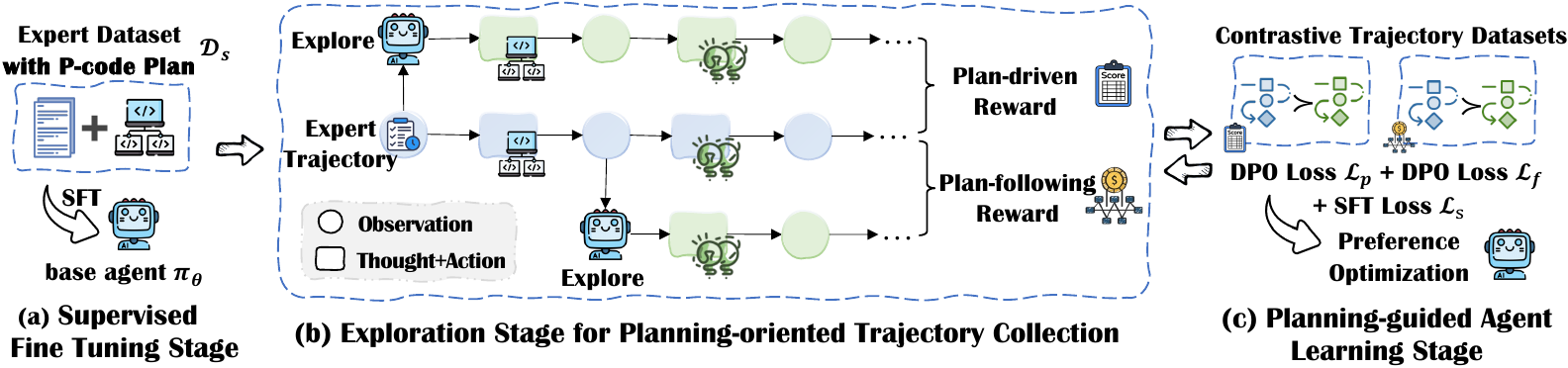}
    \caption{The overview of PGPO. Our algorithm starts by build a SFT-based agent. Then, the base agent iteratively performs exploration on expert trajectories to construct contrastive trajectory datasets based on two designed planning-oriented rewards and updates itself via preference optimization to enhance agent capabilities. }
    \label{fig:pgpo}
\end{figure*}

\subsection{P-code Plan Improves Generalization}\label{sec:plan_advantage}
We adopt supervised fine-tuning (SFT) to equip base LLMs with agent abilities and then investigate whether LLM agents can benefit from our designed pseudocode-style plans. 
Our experiments are based on four popular LLMs: Llama-2-7B/13B~\cite{touvron2023llama}, Llama-3-8B~\cite{dubey2024llama} and Mistral-7B-v0.3~\cite{jiang2023mistral}. 
For the agent tasks, we choose two representative datasets: ALFWorld~\cite{shridhar2021alfworld} and WebShop~\cite{yao2022webshop}. 
We employ the average reward as metric to reflect the agent performance. 
Note that ALFWorld comprises both seen and unseen test sets in order to evaluate the in-distribution and out-of-distribution generalization of the agents.
Please ref to Appendix~\ref{app:dataset} and~\ref{app:sft} for more details. 

We first collect ReAct-style training expert trajectories following~\citet{xiong2024ipr} and then use the above plan generation pipeline (Section~\ref{sec:plan_gen}) to obtain P-code Plans, along with natural language plans. 
These two type generated plans are finally incorporated into the first step of original trajectories to construct new training datasets, respectively. 
In Appendix~\ref{app:sft}, we present more details of dataset construction procedure with some examples.

As shown in Figure~\ref{fig:plan_based_sft}, with the help of P-code Plan, LLM agents generally have higher average reward (10 out of 12 scenarios). 
Compared to naive expert trajectories (\textit{w/o Plan}), integrating plans into training data (regardless of format) empowers LLM agents with basic planning ability, which is beneficial to the following agent reasoning. 
Regarding planning format, although \textit{w/ P-code Plan} brings weaker performance than \textit{w/ NL Plan} for Llama-2-7B and Llama-3-8B on the seen ALFWorld tasks (2 out of 12 scenarios), it consistently achieves better generalization to unseen ALFWorld tasks for all four models. 
We consider this can be attributed that abstract planning steps in P-code Plan capture generalizable meta-knowledge guiding task solution but natural language plans are prone to overfitting. 
We also analyze whether executable code format could lead to higher performance (in Appendix~\ref{app:execode}) and find verbose plans pose a greater challenge for generation, thereby negatively affect reasoning.

Moreover, we further calculate the average number of interaction turns on all evaluated instances. In Table~\ref{tab:avg_turn}, LLM agents \textit{w/ P-code Plan} surprisingly reduce the average interaction turns compared to the other two settings. 
This underscores the superiority of P-code Plan in preventing blind exploration when dealing with agent tasks. 
Another interesting finding is handling unseen tasks requires more interactions than seen tasks, but the increase from LLM agents \textit{w/ P-code Plan} is relatively low. 
This phenomenon also demonstrates our designed P-code Plan is suitable for agent generalization.

Overall, the advantage of P-code Plans over NL plans can be summarized into two aspects: (1) abstract pseudocode format helps LLM agents better generalize to unseen tasks; (2) concise and structured pseudocode facilitates a more efficient agent reasoning process with fewer interactions.
\section{P-code Plan-Guided Agent Learning}
Despite widespread use in open LLM agents~\cite{chen2023fireact,zeng-etal-2024-agenttuning,yin-etal-2024-agent}, supervised fine-tuning approach has its drawback, that is, the limited generalization ability due to overreliance on expert trajectories~\cite{song-etal-2024-trial,fu2025agentrefine}. 
Recent works focus on Direct Preference Optimization (DPO)~\cite{rafailov2024direct} and its variants to develop LLM agents with contrastive trajectory dataset~\cite{yang2024embodied,song-etal-2024-trial,xiong2024ipr,shi2024direct}. 
Inspired by our findings in Section~\ref{sec:plan_advantage}, we propose a method for improving agents' pseudocode-style planning ability, called PGPO, standing for \underline{\textbf{P}}lanning \underline{\textbf{G}}uided \underline{\textbf{P}}reference \underline{\textbf{O}}ptimization. 
The overview of our method is depicted in Figure~\ref{fig:pgpo}.

\subsection{Planning-oriented Trajectory Collection}
We start off training the base agent $\pi_{\theta}$ via SFT on the expert dataset $\mathcal{D}_s = \{(u, p, \tau)^{(i)}\}^{|\mathcal{D}_s|}_{i=1}$ with P-code Plans $p$ generated in Section~\ref{sec:plan_gen}. 
We denote new plan-incorporated trajectory as $\tau^{'}=(p, t_{1}, a_{1}, $ $o_{1},..., o_{n-1}, t_{n}, a_{n}, o_{n})$ and the loss function can be formulated as:
\begin{align}\label{eq:sft_loss}
&\mathcal{L}_{SFT}(\theta)=-\mathbb{E}_{(u,\tau^{'})\sim \mathcal{D}_s}\{\sum_{j=1}^n\left[\log{\pi_{\theta}(t_j|u,p,\tau_{j-1})}\right.\notag
\\&+\left.\log{\pi_{\theta}(a_j|u,p,\tau_{j-1},t_j)}\right]+\log{\pi_{\theta}(p|u)}\},
\end{align}
where $\tau_{j-1}=(t_{1}, a_{1}, $ $o_{1},..., a_{j-1}, o_{j-1})$ represents the interaction history of previous $j\!-\!1$ rounds. 

The obtained base agent $\pi_{\theta_{base}}$ is used for exploration on the expert trajectory to collect contrastive action pairs. 
Oriented towards planning, we design two rewards for contrastive trajectory construction: 1) plan-driven reward; 2) plan-following reward. 

\paragraph{Plan-driven Reward $r_d$} evaluates the influence of P-code Plans on the entire trajectory. 
Given each task instruction $u$, we use base agent to generate the P-code Plan $\hat{p}\sim\pi_{\theta_{base}}(\cdot|u)$ and subsequent reasoning steps $\hat{\tau}\sim\pi_{\theta_{base}}(\cdot|u,\hat{p})$. 
In our experiments, agent stops exploration when task completes or the maximum number of interaction rounds is exceeded.
Then, the environment will give the outcome reward $r_o$, which is positively correlated with the quality of exploration trajectory.
For simplicity, we use this outcome reward as $r_d$. 
By comparing $r_d(p, \tau)$ and $r_d(\hat{p}, \hat{\tau})$ for expert trajectory and exploration trajectory, we get our first contrastive trajectory dataset $D_p = \left\{(u,p^w,\tau^w,p^l,\tau^l)^{(i)}\right\}_{i=1}^{|D_p|}$. We use $(p^w,\tau^w)\succ(p^l,\tau^l)\;|\;u$ to represent the situation where $(p^w,\tau^w)$ with higher reward is preferred over $(p^l,\tau^l)$ with lower reward.

\paragraph{Plan-following Reward $r_f$} 
assesses the extent to which agents comprehend the structured intentions behind P-code Plans and their proficiency in following these plans during task execution. 
Given the first expert interaction round $(u,p,\tau_1)$ as historical trajectory, we query base agent to continue exploration $\hat{\tau}_{2:m}\sim\pi_{\theta_{base}}(\cdot|u,p,\tau_1)$. $m$ denotes the total interaction rounds. 
Here, we select the first two interaction rounds $\hat{\tau_{2}}$ as representative to analyze the alignment between plans and executed actions.
Inspired by Monto Carlo Tree Search~\cite{kocsis2006bandit}, we quantify it as the potential to complete task successfully. 
It is intuitive that when the quality of P-code Plan is guaranteed, subsequent reasoning steps that closely adhere to the plan are likely to yield higher outcome rewards. 
Therefore, we use one scorer agent to generate new subsequent trajectory $\tau^{s}_{3:m
'}\sim\pi_{\theta_{scorer}}(\cdot|u,p,\hat{\tau_{2}})$.  
By sampling $N$ new trajectories, the average outcome reward $r_o$ from the environment estimates the plan-following reward:
\begin{equation}\label{eq:plan_following_reward}
    r_f(u,p,\hat{\tau}_2) = \frac{\sum_{i=1}^{N} r_o(\tau^s_{3:m'}|u,p,\hat{\tau}_2)^{(i)}}{N}
\end{equation}
In our approach, we use $\pi_{\theta_{base}}$ as $\pi_{\theta_{scorer}}$. 
Then, similar to the construction of $D_p$, we contrast subsequent trajectory $\tau^{w}_{2:n}\succ\tau^{l}_{2:m}\;|\;(u,p,\tau_1)$ based on $r_f(u,p,\tau_2)$ and $r_f(u,p,\hat{\tau}_2)$, establishing our second contrastive trajectory dataset $D_f=\left\{(u,p,\tau_1,\tau^{w}_{2:n},\tau^{l}_{2:m})^{(i)}\right\}_{i=1}^{|D_f|}$.

\subsection{Planning-guided Agent Learning}
After collecting preference data, DPO method is utilized to optimize our base agent. 
First, based on dataset $D_p$, agent learns to generate high-quality P-code Plans along with subsequent reasoning by minimizing the following loss:
\begin{align}\label{eq:L_p}
    \mathcal{L}_{p}\!=&\!-\mathbb{E}_{\substack{(u, p^w, \tau^w, p^l, \tau^l)\\\sim\mathcal{D}_p}}\!\notag\!\left[\log\sigma\!\left(\!\beta\log\frac{\pi_{\theta}(p^w,\tau^w|u)}{\pi_{ref}(p^w,\tau^w|u)}\right.\right.\\&
    \qquad\qquad\quad\;-\!\!\left.\left.\beta\log\frac{\pi_{\theta}(p^l,\tau^l|u)}{\pi_{ref}(p^l,\tau^l|u)}\right)\right] 
\end{align}
where $\sigma$ denotes the logistic function, $\beta$ controls the weight of the preference for the reference model $\pi_{ref}$. 
Meanwhile, agent refines its parameters to develop the plan-following ability gathered from dataset $D_f$, which can be formulated as:

\vspace{-8pt}
{\mysize
\begin{align}\label{eq:L_f}
    \mathcal{L}_{f}\!=&\!-\!\mathbb{E}_{\substack{(u,p,\tau_1,\tau^w_{2:n},\tau^l_{2:m})\\\sim\mathcal{D}_f}}\!\!\left[\log\sigma\!\left(\!\beta\log\!\frac{\pi_{\theta}(\tau^w_{2:n}|u,p,\tau_1)}{\pi_{ref}(\tau^w_{2:n}|u,p,\tau_1)}\right.\right.\notag\\&
    \qquad\qquad\quad-\!\!\left.\left.\beta\log\frac{\pi_{\theta}(\tau^l_{2:m}|u,p,\tau_1)}{\pi_{ref}(\tau^l_{2:m}|u,p,\tau_1)}\right)\right]
\end{align} 
}

One issue of standard DPO is that log probability of chosen trajectories may decrease over training steps, leading to sub-optimal performance. 
Following previous works~\cite{pang2024iterative,xiong2024ipr}, we add SFT loss to mitigate this issue:
\begin{equation}\label{eq:L_s}
   \mathcal{L}_{s}\!=\!-\mathbb{E}_{(u, p^w, \tau^w, p^l, \tau^l)\sim \mathcal{D}_p} \left[\log{\pi_{\theta}(p^w,\tau^w|u)}\right] 
\end{equation}

Finally, the optimization objective of PGPO is:
\begin{equation}
\min_{\pi_{\theta}}\left(\mathcal{L}_{p}+\mathcal{L}_{f}+\mathcal{L}_{s}\right)
\end{equation}
The updated agent will be used as new base agent for exploration and iterate the above learning process until exceeding the maximum iterations. 
The overall procedure of PGPO is summarized in Appendix~\ref{app:alg} Algorithm~\ref{alg:pgpo}.

\begin{table*}[tbp]
  \centering
\resizebox{\linewidth}{!}{
  \begin{tabular}{ccccccccccc}
  \toprule
    \multirow{3}{*}{\textbf{Method}} & \multicolumn{5}{c}{\textbf{Llama-2-7B}}&\multicolumn{5}{c}{\textbf{Llama-2-13B}}\\
    \cmidrule(lr){2-6}
    \cmidrule(lr){7-11}
    &\multicolumn{2}{c}{\textbf{ALFWorld}} & \multirow{2}{*}{\textbf{WebShop}}&\multirow{2}{*}{\textbf{TextCraft}}&\multirow{2}{*}{\textbf{Avg.\;}}&\multicolumn{2}{c}{\textbf{ALFWorld}} & \multirow{2}{*}{\textbf{WebShop}}&\multirow{2}{*}{\textbf{TextCraft}}&\multirow{2}{*}{\textbf{Avg.}}\\
    &Seen&UnSeen&&&&Seen&UnSeen&&&\\
    \midrule
    SFT&60.0&67.2&60.2&28.0&53.9\;&67.1&67.9&62.2&29.0&56.6\\
    ETO&68.6&72.4&67.4&\underline{35.0}&60.9\;&75.0&69.4&68.9&\underline{42.0}&63.8\\
    IPR&\underline{70.3}&\underline{74.7}&\underline{71.3}&34.0&\underline{62.6}\;&75.0&\underline{76.9}&\underline{72.2}&39.0&\underline{65.8}\\
    PGPO&\textbf{76.4}&\textbf{76.9}&\textbf{72.2}&\textbf{43.0}&\textbf{67.1}\;&\textbf{77.1}&\textbf{77.6}&\textbf{73.7}&\textbf{48.0}&\textbf{69.1}\\
    \midrule[1pt]
    \multirow{3}{*}{\textbf{Method}} &\multicolumn{5}{c}{\textbf{Llama-3-8B}}&\multicolumn{5}{c}{\textbf{Mistral-7B}}\\
    \cmidrule(lr){2-6}\cmidrule(lr){7-11}
    &\multicolumn{2}{c}{\textbf{ALFWorld}} & \multirow{2}{*}{\textbf{WebShop}}&\multirow{2}{*}{\textbf{TextCraft}}&\multirow{2}{*}{\textbf{Avg.\;}}&\multicolumn{2}{c}{\textbf{ALFWorld}} & \multirow{2}{*}{\textbf{WebShop}}&\multirow{2}{*}{\textbf{TextCraft}}&\multirow{2}{*}{\textbf{Avg.}}\\
    &Seen&UnSeen&&&&Seen&UnSeen&&&\\
    \midrule
    SFT&67.1&72.4&61.2&20.0&55.2\;&72.1&68.7&61.8&31.0&58.4\\
    ETO&72.1&73.1&66.2&36.0&61.9\;&\textbf{75.0}&72.4&66.2&\underline{38.0}&62.9\\
    IPR&\underline{72.9}&\underline{73.9}&\underline{72.0}&\underline{38.0}&\underline{64.2}\;&\underline{73.6}&\underline{73.1}&\textbf{69.6}&36.0&\underline{63.1}\\
    PGPO&\textbf{75.0}&\textbf{76.9}&\textbf{72.3}&\textbf{46.0}&\textbf{67.6}\;&\textbf{75.0}&\textbf{77.6}&\underline{69.0}&\textbf{45.0}&\textbf{66.7}\\
  \bottomrule
  \end{tabular}
}
  \caption{\label{tab:main_result}
    Main results of PGPO compared to training-based baselines on ALFWorld, WebShop and TextCraft. \textbf{Bold} and \underline{underline} indicate the best and the second-best results of each model. For all methods (except SFT), we report the best performance across all iterations following~\citet{xiong2024ipr}. Our PGPO is evaluated in zero-shot setting. 
  }
\end{table*}

\section{Experiments}
\subsection{Experimental Settings}
\paragraph{Datasets and Metrics.} Besides \textbf{ALFWorld} for embodied household tasks and \textbf{WebShop} for online shopping (as described in Section~\ref{sec:plan_advantage}), we also include one game benchmark \textbf{TextCraft}~\cite{prasad2024adapt} for crafting Minecraft items. 
ALFWorld and TexCraft provide binary rewards to indicate task success while WebShop provides dense rewards from 0 to 1 to measure the task completion level. 
For all datasets, we choose the average reward as evaluation metric. 
Detailed statistical information of the datasets are in Appendix~\ref{app:dataset}.

\begin{table}[tbp]
\setlength\tabcolsep{3pt}
\centering
\resizebox{\linewidth}{!}{
  \begin{tabular}{lcccc}
  \toprule
    \multirow{2}{*}{\textbf{Method}}&\multicolumn{2}{c}{\textbf{ALFWorld}} &\multirow{2}{*}{\textbf{WebShop}}&\multirow{2}{*}{\textbf{TextCraft}}\\\
    &Seen&UnSeen&&\\
    \midrule
    ReAct\small{+GPT-4}&42.9&38.1&63.2&28.0\\
    ReAct\small{+GPT-3.5}&7.9&10.5&62.4&20.0\\
    ADaPT\small{+GPT-4}&\underline{75.0}&69.4&64.8&\textbf{48.0}\\
    ADaPT\small{+GPT-3.5}&70.3&\underline{71.6}&62.7&26.0\\
    \midrule
    \textbf{PGPO}\small{+Llama-2-7B}&\textbf{76.4}&\textbf{76.9}&\underline{72.2}&43.0\\
    \textbf{PGPO}\small{+Llama-3-8B}&\underline{75.0}&\textbf{76.9}&\textbf{72.3}&\underline{46.0}\\
  \bottomrule
  \end{tabular}
}
  \setlength{\belowcaptionskip}{-10pt}
  \caption{\label{tab:prompt_baseline}Comparative experiments on PGPO vs. prompt-based baselines. The best and second-best results are marked in \textbf{bold} and \underline{underline}.}
\end{table}

\paragraph{Baselines.} We compare PGPO with naive SFT and other two leading agent learning methods: (1) ETO~\cite{song-etal-2024-trial} applies DPO loss to improve the agent from its exploration failures; (2) IPR~\cite{xiong2024ipr} introduces step-wise process supervision into LLM agent training. 
For fair comparison, we also select Llama-2-7B/13B, Llama-3-8B and Mistral-7B-v0.1 as backbone models and  use the same training data. 
Additionally, we include two strong closed-source LLMs: GPT-3.5-Turbo and GPT-4, utilizing two prompt-based methods ReAct~\cite{yao2023react} and ADaPT~\cite{prasad2024adapt} for comparison.

\vspace{-5pt}
\paragraph{Implementation Details.} During the SFT phase, we set the learning rate as 2e-5 and the batch size as 48 across 3 training epoches. The cosine scheduler is employed with a 3\% warm-up ratio. 
In the trajectory collection stage, the base agent explores the environment using a temperature of 0. To construct contrastive pairs, we sample $N$=5 times with temperature=1 to calculate the plan-following reward. 
During the following preference optimization phase, we tune the learning rate from 5e-7 to 5e-6 and test two values for $\beta$ in the DPO loss: 0.01 and 0.1. 
The maximum number of iterations is set to 4. 
All training experiments are conducted on 8 NVIDIA A100 80 GB GPUs.
See Appendix~\ref{app:sft}-~\ref{app:eval} for more details.

\begin{table}[tbp]
\setlength\tabcolsep{4pt}
\centering
\resizebox{\linewidth}{!}{
  \begin{tabular}{ccccc}
  \toprule
    &\multicolumn{2}{c}{\textbf{ALFWorld}} & \multirow{2}{*}{\textbf{WebShop}}&\multirow{2}{*}{\textbf{TextCraft}}\\
    &Seen&UnSeen&&\\
    \midrule
    \textbf{PGPO}&76.4&76.9&72.2&43.0\\
    \textit{- P-code}&75.7\small{\textcolor{darkblue}{$\downarrow0.7$}}&71.6\small{\textcolor{darkblue}{$\downarrow5.3$}}&69.6\small{\textcolor{darkblue}{$\downarrow2.6$}}&40.0\small{\textcolor{darkblue}{$\downarrow3.0$}}\\
    \textit{- $L_f$}&74.3\small{\textcolor{darkblue}{$\downarrow2.1$}}&75.4\small{\textcolor{darkblue}{$\downarrow1.5$}}&70.4\small{\textcolor{darkblue}{$\downarrow1.8$}}&41.0\small{\textcolor{darkblue}{$\downarrow2.0$}}\\
    \textit{- $L_s$}&69.3\small{\textcolor{darkblue}{$\downarrow7.1$}}&68.7\small{\textcolor{darkblue}{$\downarrow8.2$}}&64.8\small{\textcolor{darkblue}{$\downarrow7.4$}}&35.0\small{\textcolor{darkblue}{$\downarrow8.0$}}\\
  \bottomrule
  \end{tabular}
}
  \setlength{\belowcaptionskip}{-10pt}
  \caption{\label{tab:ablation}Approach ablations of PGPO. Experiments are based on Llama-2-7B. \textit{- P-code} represents using NL plans to replace P-code Plans. \textit{- $L_f$} denotes leaving out the estimation of plan-following reward $r_f$, followed by the removal of $L_f$. \textit{- $L_s$} means the removal of SFT loss.}
\end{table}

\subsection{Main Results}
Table~\ref{tab:main_result} and~\ref{tab:prompt_baseline} show the evaluation results of PGPO and baselines on three agent benchmarks.
First, compared with training-based baselines, PGPO significantly increases the average reward across all the datasets. 
Specifically, PGPO with Llama-2-7B surpasses the state-of-the-art method IPR by an improvement of 7.2\% on average reward. 
This indicates the incorporation of P-code Plans into training data can provide the model with enhanced reasoning abilities in accomplishing the agent tasks.

Second, for prompt-based baselines, traditional ReAct paradigm using GPT-3.5-Turbo exposes poor performance on all agent datasets. 
Although ADaPT+GPT-3.5 gains a performance boost via recursive task decomposition, it still underperforms our method. 
In particular, PGPO+Llama-2-7B surpasses ADaPT+GPT-3.5 by relative margins of 9.5\% and 17\% points on WebShop and TextCraft.
While prompting methods with GPT-4 improve the agent performance, PGPO+Llama-3-8B alleviates the need for few-shot context and achieves comparable or even better results. 
This demonstrates smaller open-source models can be effective agents through iterative training, rivaling or exceeding the agent capabilities of strong closed-source models.

\begin{figure}[tbp]
    \centering
    \includegraphics[width=1\linewidth]{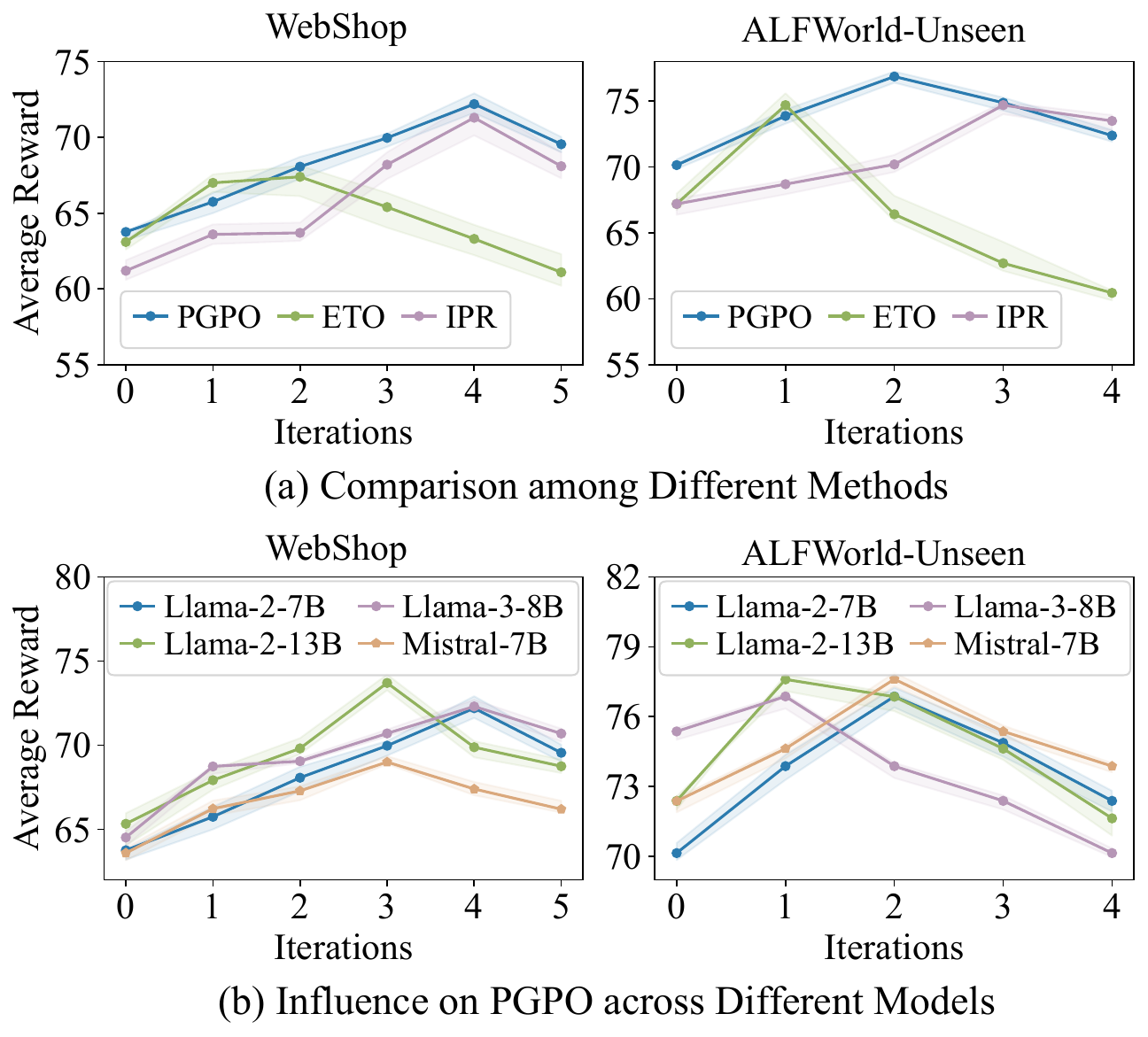}
    \setlength{\abovecaptionskip}{-8pt}
    \setlength{\belowcaptionskip}{-10pt}
    \caption{Ablation study on optimization iterations. (a) provides a comparison of the performance of PGPO against ETO and IPR across varying iterations. (b) shows the influence of increasing iterations on PGPO across different base models. iter=0 is the SFT stage. Shaded regions indicate standard error across 5 trails.
    \label{fig:ablation_on_iteration}} 
\end{figure}

Finally, we focus on the effectiveness of PGPO across different base models on various datasets. 
(1) Generalization on different models: Besides the LLaMA family of models, we also include Mistral-7B in Table~\ref{tab:main_result} and Qwen2.5 series in Appendix~\ref{app:qwen}. Regardless of model sizes and families, our method consistently exhibit the advantage of P-code Plans guiding agent reasoning. 
(2) Generalization on diverse unseen tasks: To comprehensively assess its capability in out-of-distribution scenarios, we conduct additional experiments on ScienceWorld~\cite{wang2022scienceworld} in Appendix~\ref{app:sciworld} Table~\ref{tab:sciworld}. 
PGPO outperforms the ETO and IPR baselines in generalizing to unseen tasks, where the performance gap is larger when dealing with more complex interactive tasks. 


\begin{figure}[tbp]
    \centering
    \includegraphics[width=0.98\linewidth]{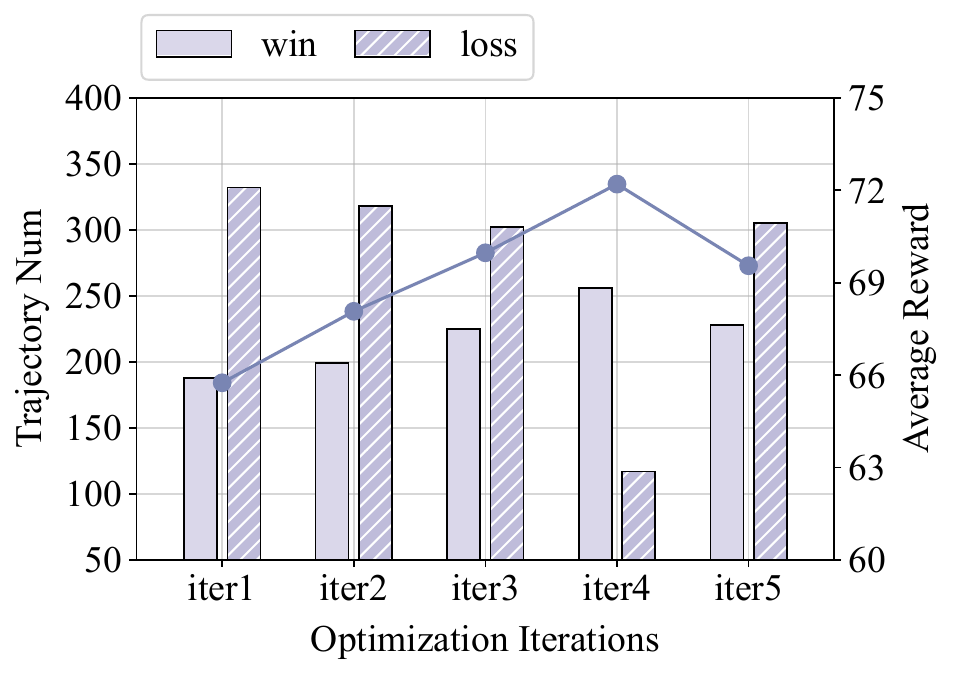}
    \setlength{\abovecaptionskip}{-8pt}
    \setlength{\belowcaptionskip}{-10pt}
    \caption{The correlation between collected contrastive trajectory dataset distribution and agent performance during iterative optimization. Here, "win" means agent-generated trajectory surpassing expert trajectory while "loss" represents falling short.}
    \label{fig:win_loss}
\end{figure}

\begin{figure}[tbp]
    \centering
    \includegraphics[width=0.7\linewidth]{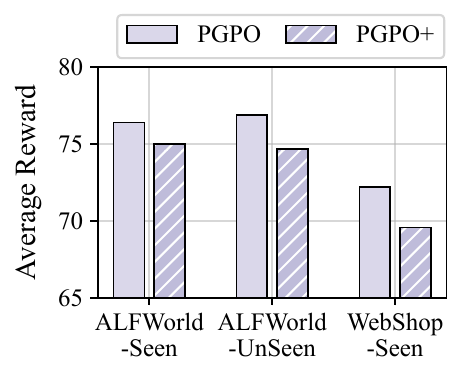}
     \setlength{\abovecaptionskip}{5pt}
     \setlength{\belowcaptionskip}{-10pt}
    \caption{\label{fig:step_wise}Comparison between PGPO+ and PGPO.}
\end{figure}

\subsection{Ablation Study}~\label{sec:ablation}
\vspace{-20pt}
\paragraph{Approach Ablations.} 
Table~\ref{tab:ablation} illustrates that the performance of PGPO obviously declines after removing certain key components. 
We observe that the most significant performance drop comes from the removal of SFT loss (\textit{- $L_s$}), which is consistent with previous findings~\cite{pang2024iterative}. 
To demonstrate the advantage of P-code Plans over NL plans, we specifically replace P-code Plans with NL plans (generation pipeline is same with Section~\ref{sec:plan_advantage}) in training data and then employ PGPO. 
As expected, the results indicate even with subsequent iterative preference optimization, the performance upper bound brought by P-code Plans is still superior to that of NL plans.
Additionally, the effect of using plan-following reward to include $L_f$ in optimization loss is better than \textit{- $L_f$}, indicating the necessity of plan-following rewards.

\vspace{-5pt}
\paragraph{Ablation on Optimization Iterations.}
Figure~\ref{fig:ablation_on_iteration} shows the iteration ablation results from two aspects: 
(a) With optimization iterations increase, 
all methods first exhibit performance improvements and then deteriorate due to excessive iterations. 
Among them, PGPO consistently achieves the highest peak performance. 
We consider that this can be attributed to the excellent starting point (i.e., iter=0) in PGPO since the incorporation of P-code Plans into SFT data effectively enhances the agent reasoning capability. 
(b) Despite different base models, the peak performance of PGPO can be achieved within 4 iterations. 
However, the performance variation trend across them quite differs, which reflects the distinct extent of each model grasping the meta-knowledge inherent in 
P-code Plans. 
Furthermore, as depicted in Figure~\ref{fig:win_loss}, agent achieves optimal performance during iterative optimization when the number of its exploration trajectory surpassing expert trajectory reaches a peak. 

\begin{figure*}[tbp]
    \includegraphics[width=1\linewidth]{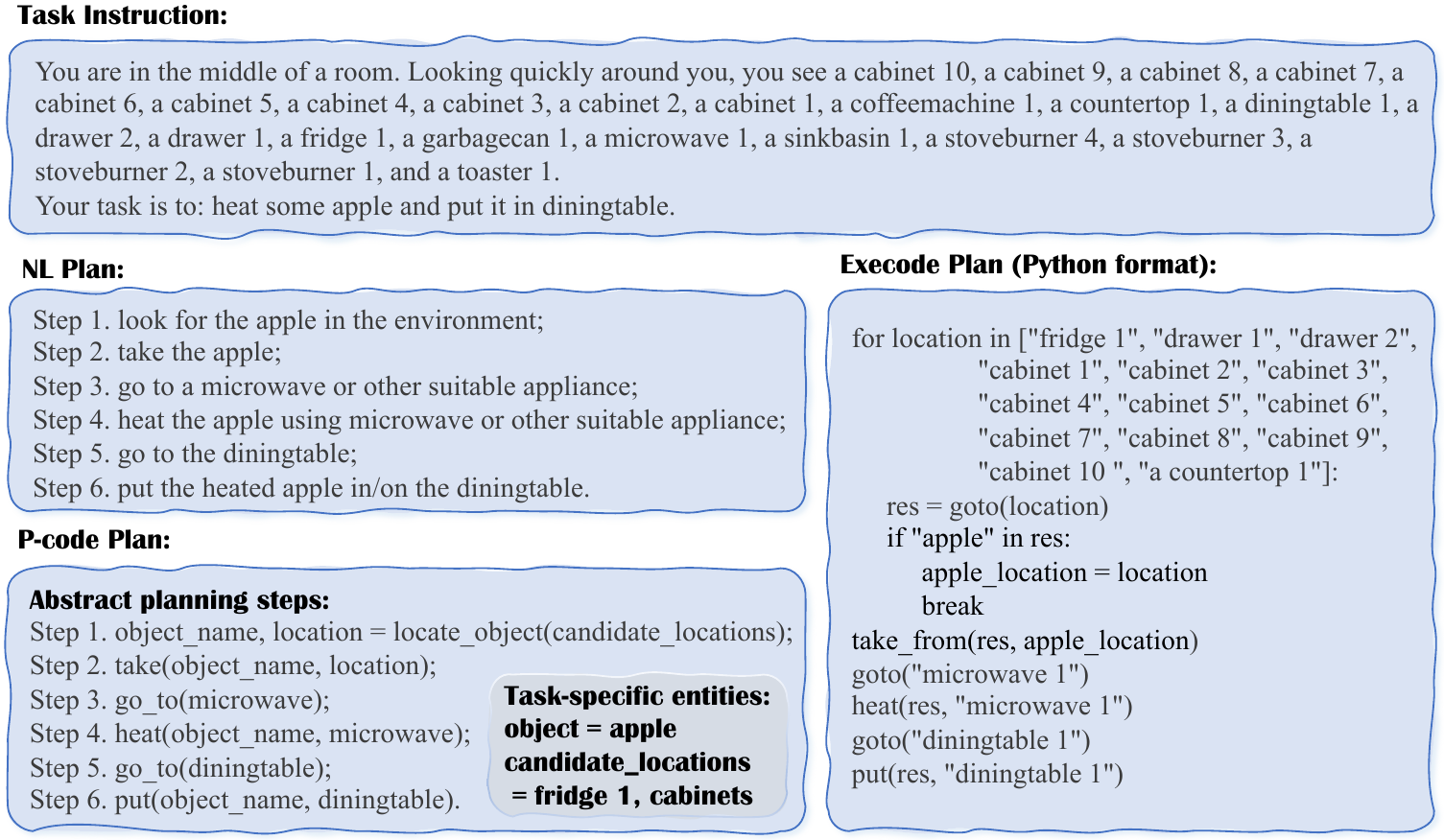}
    \setlength{\abovecaptionskip}{-5pt}
    \setlength{\belowcaptionskip}{-5pt}
    \caption{Case study for our P-code Plan compared with other formats.}
    \label{fig:plan_example}
\end{figure*}

\begin{table}[tbp]
    \centering
    \begin{minipage}[b]{0.52\linewidth} 
    \centering
    \resizebox{\linewidth}{!}{
    \begin{tabular}{ccc}
        \toprule
        \multirow{2}{*}{\textbf{Method}}&\multicolumn{2}{c}{\textbf{ALFWorld}} \\
        &Seen&UnSeen\\
        \midrule
        ETO&34.28\%& 32.83\%\\
        IPR&30.71\%& 28.35\%\\
        PGPO&\textbf{23.57\%}&\textbf{26.86\%}\\
        \toprule
    \end{tabular}}
    \setlength{\abovecaptionskip}{5pt}
    \setlength{\belowcaptionskip}{-10pt}
    \caption{Invalid action rate on ALFWorld.}
    \label{tab:invalid_action}
    \end{minipage}
    \hspace{0.05\linewidth} 
    \begin{minipage}[b]{0.38\linewidth} 
    \centering
    \resizebox{\linewidth}{!}{
    \begin{tabular}{cc}
        \toprule
        \textbf{Method}&\textbf{Webshop} \\
        \midrule
        ETO& 37.5 \\
        IPR& 40.5\\
        PGPO& \textbf{41.0}\\
        \toprule
    \end{tabular}}
    \setlength{\abovecaptionskip}{5pt}
    \setlength{\belowcaptionskip}{-10pt}
    \caption{Success rate on WebShop.}
    \label{tab:success_rate}
    \end{minipage}%
\end{table}

\subsection{Analysis}
\paragraph{Analysis on training time efficiency.} 
We compare the time consumption of PGPO with two training-based baselines on WebShop. 
Under the same resource constraints, ETO, IPR and PGPO first undergo a 1-hour SFT phase and additionally require 1.5h, 4.5h and 3.2h per optimization iteration, respectively. 
Therefore, PGPO delivers a 9\% performance improvement while maintains reasonable training efficiency, requiring less than twice the time cost of ETO. 

\vspace{-5pt}
\paragraph{P-code Plan guidance can reduce the incidence of action errors and omissions in reasoning.} 
Taking ALFWorld as an example, we calculate the proportion of trajectories containing invalid actions for each method on Llama-2-7B in Table~\ref{tab:invalid_action}. 
The results demonstrate PGPO decreases action errors with the help of P-code Plans. 
Then we analyze the action omissions of agents via the task success rate. 
Since WebShop provides dense rewards, the trajectory is considered success only when final reward is 1.0, i.e., agent has selected all necessary product attributes without any omissions. 
From Table~\ref{tab:success_rate}, it can be observed that PGPO achieves the highest success rate, indicating guidance from P-code Plans indeed reduces the agent's action omissions.

\vspace{-5pt}
\paragraph{Step-wise reward does not necessarily elicit better LLM agents.} 
Regarding the design of plan-following reward in PGPO, we only consider the alignment between the first step (containing the generated plans) and the second step to construct contrastive trajectory pairs. 
Since step-level process supervision has been effectively utilized in reasoning tasks~\cite{lightman2024lets}, 
we evaluate whether introducing step-wise reward to our PGPO could further facilitate agent reasoning. 
Following IPR~\cite{xiong2024ipr}, we add step-level rewards into our method (denoted PGPO+). 
It can be observed from Figure~\ref{fig:step_wise} that step-wise reward negatively impacts the PGPO performance. 
To speculate on the reason behind, we manually check the quality of training data collected by PGPO+ and find that although step-level process supervision increases the data scale, some constructed preference pairs may be ambiguous, which poses a potential risk of reward hacking~\cite{gao2023scaling}. 
Therefore, it is still challenging to accurately determine the contribution of the intermediate step, thus introducing step-wise reward instead plays a negative role in agent reasoning~\cite{guo2025deepseek}.


\subsection{Case Study}
In Figure~\ref{fig:plan_example}, we show the generated P-code Plan compared with other two plan formats within the same task in ALFWorld. 
First, NL plans are less structured than plans in pseudocode or executable code format 
since elements such as articles and conjunctions may have no role in complex reasoning logic.
Second, as described in Appendix~\ref{app:execode}, Execode Plan is more verbose than P-code Plan. 
In this case, the Execode Plan lists almost all of locations 
for agent to explore, guaranteeing the solvability of this agent task. 
However, this may introduce unnecessary context and lead to blind trial-and-error, resulting in task failure due to exceeding the maximum interaction turns. 
By contrast, our P-code Plan strikes a balance between structural rigor and concision, thereby facilitating agent reasoning.

\section{Related Works}
\paragraph{LLM Agents.} 
The remarkable capabilities of LLMs have spurred research into developing AI agents. 
These LLM agents are generally equipped with reasoning and acting capabilities, enabling them to handle a wide range of tasks~\cite{autogpt,babyagi,liu2024agentlite}. 
Prompt-based methods like ReAct~\cite{yao2023react}, Reflexion~\cite{shinn2024reflexion} and ADaPT~\cite{prasad2024adapt} utilize strong closed-source LLMs to build powerful agents.
However, prompting strategies are heavily dependent on those enhanced but expensive LLMs, resulting in high usage costs.
Recent studies explore the fine-tuning methods based on open-source LLMs to improve agent intelligence~\cite{chen2023fireact,zeng-etal-2024-agenttuning}.

\paragraph{Agent Planning.} 
Planning plays a crucial role in agent reasoning, with different planning paradigms, such as Plan\&Solve~\cite{wang2023plan} and PlanAct~\cite{liu2023bolaa}, offering diverse approaches to agent tasks. 
Few works have explored the potential of alternative plan formats beyond natural language. 
To utilize the precision of formal language, ~\citet{li2024formal} constructs a context-free grammar to control the NL plan generation. 
~\citet{zhang2024lamma} rely on translating NL tasks into  Planning Domain Definition Language (PDDL) and then solve problems with PDDL planners. 
~\citet{silver2024generalized} consider PDDL domains and directly use LLMs for generalized planning. 
However, the above studies just use such structured language to assist agent planning rather than empower LLMs to generate structured plans for enhanced reasoning.
Concurrently with our work, ~\citet{wen2025unlocking} introduce code-form plans to do reasoning tasks under the few-shot setting without fine-tuning.

\paragraph{Agent Learning.} 
Previous works focus on learning from expert trajectory data to align agent behavior with expert~\cite{chen-etal-2024-agent,yin-etal-2024-agent}. 
Recently, learning from preference has shown promise for developing LLM agents. 
NAT~\cite{wang2024learning} teaches the model to
differentiate between correct and incorrect interactions during fine-tuning. 
ETO~\cite{song-etal-2024-trial} leverages iterative explored trajectories for training via DPO loss. 
IPR~\cite{xiong2024ipr} constructs step-wise contrastive action pairs using estimated step-level rewards to guide the agent optimization process.
These efforts highlight iterative preference learning techniques unlock sophisticated agent capabilities. 

\section{Conclusion}
In this paper, we propose PGPO, which empowers LLM agents with enhanced reasoning capabilities under the guidance of pseudocode-style plans. 
Our motivation is based on that abstract P-code Plan can capture efficient structural logic of reasoning compared with NL plans, suitable for LLM agent's generalization to analogous agent tasks.
After incorporating automatically generated P-code Plans into existing ReAct-style datasets, PGPO starts by a competent base agent through SFT.
Then, PGPO iteratively refines the base agent via preference optimization based on two planning-oriented rewards. 
Experimental results demonstrate PGPO effectively achieves new SOTA performance across three agent benchmarks. 
Further analysis shows that P-code Plan exhibits robust potential in mitigating action errors and critical step omissions during reasoning.

\section*{Limitations}
This paper focuses on incorporating pseudocode-style plans to guide agent preference optimization. 
Despite its new SOTA performance, we acknowledge the following limitations of our work: 
1) Our method deploys Monte Carlo sampling to estimate plan-following reward, incurring additional inference costs compared to the ETO baseline. However, sampling only needs to be conducted for one step per iteration, which is more efficient than the design of step-wise reward in IPR baseline. And ablation studies in Section~\ref{sec:ablation} demonstrates the necessity of plan-following rewards.
2) Our method designs structured P-code Plan to enhance agent reasoning. Although powerful GPT-4o guarantees the plan quality to a certain extent, it is still necessary to research how to verify plans automatically beyond human verification.

In the future, we plan to conduct research on rule-based rewards~\cite{guo2025deepseek} since the structured nature of our P-code Plan provides an interpretable scaffold for automating rule-based reward design. 
Furthermore, we explore extending our method to a multi-task training scenario, which can contribute to more generalized LLM agents.


\bibliography{custom}

\appendix

\section{PGPO Algorithm}\label{app:alg}
We summarize the workflow of PGPO in Algorithm~\ref{alg:pgpo}. 
Our algorithm starts by a Supervised Fine Tuning (SFT) stage. 
In this stage, expert dataset with P-code Plans generated in Section~\ref{sec:plan_gen} is utilized to equip the base LLM with agent capabilities. 
Next, the algorithm proceeds with the data preparation stage. 
For each expert trajectory, the plan-following reward is calculated via Monte Carlo sampling method. 
Then, the base agent explores on expert trajectories to collect new trajectory data.
Based on two planning-oriented rewards, two contrastive trajectory datasets are constructed. 
Finally, in the preference optimization stage, the base agent refines its parameters to improve its reasoning capability. 
The above data preparation and preference optimization stage will be repeated until exceeding the maximum iterations. 

\section{Implementation Details}\label{app:experiments}

\subsection{For Datasets}\label{app:dataset}
Table~\ref{tab:datasets} summarizes the statistics information of the three agent datasets. 
For ALFWorld and WebShop, we choose the ReAct-style expert trajectories collected by~\citet{xiong2024ipr}. 
For TextCraft, we primarily use the train set from AgentTraj-L~\cite{xi2024agentgym}. 
Note that unseen tasks in ALFWorld ref to new task instances with possibly known object-receptacle pairs, but always in unseen rooms with different receptacles and scene layouts than in training data.
\begin{table}[htbp]
\setlength\tabcolsep{4pt}
    \centering
    \resizebox{\linewidth}{!}{
    \begin{tabular}{cccc}
    \toprule
    \textbf{Dataset}  & \textbf{\#Train}& \textbf{\#Test}& \textbf{\#Turns}\\
    \midrule
    ALFWorld & 2851 & 274 (140-Seen, 134-Unseen)&7.97\\
    WebShop & 1624 & 200 (200-Seen) & 3.64\\
    TexrtCraft & 373 & 100 (100-Unseen) & 7.86\\
    \bottomrule
    \end{tabular}}
    \vspace{-5pt}
    \caption{Statistics information of ALFWorld, WebShop and TextCraft. "\#Turns" denotes the average number of iteraction turns for the expert trajectories.}
    \vspace{-5pt}
    \label{tab:datasets}
\end{table}

\begin{algorithm2e*}
\mysizeone
\caption{Workflow of \textbf{PGPO}}\label{alg:pgpo}
\SetInd{0.5em}{0.6em}
\KwIn{Base LLM $\pi_\theta$, Expert Dataset $\mathcal{D}_s = \{(u, p, \tau)^{(i)}\}^{|\mathcal{D}_s|}_{i=1}$: $\tau=(t_{1}, a_{1}, o_{1},$ $ ..., t_{n}, a_{n}, o_{n})$, \\Training Epoch $T_1$ for Supervised Fine Tuning, Training Epoch $T_2$ for Preference Optimization, Sampling Number $N$ for Reward Estimation, Maximum Number of Iterations $I$
}
\KwOut{The Enhanced LLM agent $\pi_\theta$}
\Comment{\textcolor[rgb]{0.25, 0.5, 0.75}{Supervised Fine Tuning Stage}}
\For{$epoch = 1$ \KwTo $T_1$}{
$\pi_\theta \leftarrow \mathop{\arg\min}\limits_{\pi_\theta}-\mathbb{E}_{\substack{(u,p,\tau)\\\sim \mathcal{D}_s}}\{\sum\limits_{j=1}^n\left[\log{\pi_{\theta}(t_j|u,p,\tau_{j-1})}\right.+\left.\log{\pi_{\theta}(a_j|u,p,\tau_{j-1},t_j)}\right]+\log{\pi_{\theta}(p|u)}\}$;\\
} 
$\pi_{scorer}=\pi_\theta$;\\
\For{$(u,p,\tau) \in \mathcal{D}_s$}{
Given the first two expert interaction rounds $\tau_2=(t_1,a_1,o_1,t_2,a_2,o_2)$ as historical trajectory, use scorer agent to sample $N$ subsequent trajectories: $\tau^{s}_{3:m'}\sim\pi_{\theta_{scorer}}(\cdot|u,p,\tau_{2})$;\\
Compute plan-following reward $r_f(u,p,\tau_2)= \frac{1}{N}\sum_{i=1}^{N} r_o(\tau^s_{3:m'}|u,p,\tau_2)^{(i)}$ of expert trajectory;\\
}
\For{$iter = 1$ \KwTo $I$}{
$\pi_{base}=\pi_\theta$; $\pi_{ref}=\pi_\theta$\\
\Comment{\textcolor[rgb]{0.25, 0.5, 0.75}{Exploration Stage for Planning-oriented Trajectory Collection}}
\For{$u \in \mathcal{D}_s$}{
Get the P-code Plan and subsequent reasoning steps from base agent: $\hat{p}\sim\pi_{\theta_{base}}(\cdot|u)$, $\hat{\tau}\sim\pi_{\theta_{base}}(\cdot|u,\hat{p})$;\\
Compare plan-driven rewards $r_d$ of $(\hat{p},\hat{\tau})$ with expert trajectory $(p,\tau)$ to get $(p^w,\tau^w)\succ(p^l,\tau^l)\;|\;u$;\\
Given the first expert interaction round $(u,p,\tau_1)$ as historical trajectory, get subsequent trajectory from base agent: $\hat{\tau}_{2:m}\sim\pi_{\theta_{base}}(\cdot|u,p,\tau_1)$;\\
Similar to line 5-6, compute plan-following reward $r_f(p,u,\hat{\tau_2})$ of agent-generated trajectory;\\
Compare $r_f(p,u,\hat{\tau_2})$ and $r_f(u,p,\tau_2)$ to get $\tau^{w}_{2:n}\succ\tau^{l}_{2:m}\;|\;(u,p,\tau_1)$;\\
}
Construct two contrastive trajectory datasets: $D_p = \left\{(u,p^w,\tau^w,p^l,\tau^l)^{(i)}\right\}_{i=1}^{|D_p|}$, $D_f=\left\{(u,p,\tau_1,\tau^{w}_{2:n},\tau^{l}_{2:m})^{(i)}\right\}_{i=1}^{|D_f|}$;\\
\Comment{\textcolor[rgb]{0.25, 0.5, 0.75}{
Planning-guided Agent Learning Stage}}
\For{$epoch = 1$ \KwTo $T_2$}{
$\pi_\theta \leftarrow \mathop{\arg\min}\limits_{\pi_\theta}\left(\mathcal{L}_{p}+\mathcal{L}_{f}+\mathcal{L}_{s}\right)$ using Eq.~\ref{eq:L_p}, Eq.~\ref{eq:L_p} and Eq.~\ref{eq:L_s};\\
}
}
\Return the enhanced LLM agent $\pi_\theta$.
\end{algorithm2e*}

\subsection{For P-code Plan Generation}\label{app:plan_gen}
We illustrate the plan generation pipeline in Figure~\ref{fig:plan_gen}. 
Based on existing ReAct-style datasets (see Appendix~\ref{app:dataset} for data source),  we first extract the agent thoughts part for preparation. 
Then, we use GPT-4o to summarize the step-by-step plan following our predefined P-code Plan format via few-shot prompting strategy. 
For each expert trajectory, only one P-code Plan is generated since we use greedy inference (temperature=0) in GPT-4o. 
One example prompt for plan distillation is shown below:

\begin{tcolorbox}[title={Example Prompt for Plan Distillation},breakable,]
Given the [Task Description], [Task] and [Solution Trajectory], you should summarize the step-by-step [Plan] in natural language for solving the task. Please note that the generated [Plan] should be global and do not contain the information from "Observation" part of [Solution Trajectory]. Then, you should format the generated [Plan] to strictly follow the pseudocode format and output in this format:\\
Step 1. ... \\
Step 2. ... \\
Step 3. ... \\
.... \\
Here is one example.\\
\textbf{<one-shot demonstration>}\\
\\
Now is your turn.\\
\;[Task Description]: \textbf{<task\_description>} \\
\;[Task]: \textbf{<task>}\\
\;[Solution Trajectory]: \textbf{<agent\_thoughts>}\\
\;[NL Plan]: \\
\;[P-code Plan]: 
\end{tcolorbox}

\subsection{For Human Verification}
To guarantee the quality of generated P-code Plans, we ask three NLP researchers to check the plan formats and their consistency with original trajectories. 
According to statistic, only nearly 15\% of the generated data need to be refined by human, while 85\% are well structured. 
This demonstrates the reliability of our automatic P-code Plan generation pipeline, enabling scalable and quality-controlled data synthesis.

\subsection{For SFT Stage}\label{app:sft}
First, after distilling P-code Plans from agent thoughts, we need to incorporate these plans into original ReAct-style datasets. 
As described in Section~\ref{sec:plan_advantage}, they are added into the first step of original trajectories.
We prefix the plans with "First, I devise a plan for solving the task:" for incorporation and list one example of new constructed training data as below:
\begin{tcolorbox}[title={SFT Data Example},breakable,]
\{\\
"from": "human",\\
"value": "\textbf{<task\_description>}"\\
\},\{\\
"from": "gpt",\\
"value": "OK"\\
\},\{\\
"from": "human",\\
"value": "\textbf{<task>}"\\
\},\{\\
"from": "gpt",\\
"value": "\textbf{Thought}: First, I devise a plan for solving the task: \textbf{<distilled P-code Plan>}\\Now, I need to first check ...\\\textbf{Action}: go to toiletpaperhanger 1"\\
\},\{\\
"from": "human",\\
"value": "\textbf{Observation}: On the toiletpaperhanger 1, you see ..."\\
\}, ...
\end{tcolorbox}
Next, we choose full-parameter fine-tuning for all models using FastChat framework. 
We detail the hyperparameters for SFT stage in Table~\ref{tab:sft_setting}.
\begin{table}[htbp]
    \centering
    \resizebox{0.8\linewidth}{!}{
    \begin{tabular}{cc}
    \toprule
    \textbf{Name} & \textbf{Value} \\
    \midrule
        num\_train\_epochs & 3\\
        train\_batch\_size & 48 \\
        per\_device\_train\_batch\_size & 4 \\
        per\_device\_eval\_batch\_size & 4 \\
        gradient\_accumulation\_steps & 2 \\
        learning\_rate & 2e-5 \\
        weight\_decay & 0. \\
        warmup\_ratio & 0.03 \\
        lr\_scheduler\_type & "cosine" \\
        model\_max\_length& 4096 \\
    \bottomrule
    \end{tabular}}
    \caption{Detailed hyperparameters used in SFT stage.}
    \label{tab:sft_setting}
\end{table}

\subsection{For Baselines}\label{app:baselines}
In this section, we provide a detailed introduction to the baselines, as well as our reproduction details. 
\begin{itemize}[leftmargin=12pt,topsep=4pt,itemsep=0pt]
    \item ETO~\cite{song-etal-2024-trial}: This framework comprises two training phases: (1) behavior cloning stage, wherein the agent undergoes fine-tuning on expert trajectory data, followed by (2) learning from failures, which employs DPO~\cite{rafailov2024direct} for subsequent policy refinement. 
    \item IPR~\cite{xiong2024ipr}: The iterative step-level process refinement framework enhances agent learning through step-by-step guidance. Via step-level reward estimation, IPR identifies discrepancies between agent-generated trajectories and the expert trajectories, thereby boosting the agent performance.
\end{itemize}
To reproduce experimental results, we maintain all the default hyperparameters in their public code\footnote{ETO: https://github.com/Yifan-Song793/ETO, \\IPR: https://github.com/WeiminXiong/IPR} and carefully extend them to TextCraft dataset.
\begin{itemize}[leftmargin=12pt,topsep=4pt,itemsep=0pt]
    \item ReAct~\cite{yao2023react}: ReAct first integrates Chain-of-Thought (CoT) into LLM agent systems through a structured Thought-Action-Observation reasoning format. For the ReAct implementation, we adopt one-shot prompting for agent reasoning.
    \item ADaPT~\cite{prasad2024adapt}: ADaPT dynamically decomposes complex tasks through recursive planning, automatically adjusting decomposition depth based on real-time feedback to align LLM competencies with evolving task demands. In our paper, we constrain the maximum interaction turns for fair comparison and directly use the open-source code for reproduction\footnote{https://github.com/archiki/ADaPT}.
\end{itemize}
Regarding other traditional prompting methods like Plan\&Solve~\cite{wang2023plan} and Reflexion~\cite{shinn2024reflexion}, we do not include them for comparison in Table~\ref{tab:prompt_baseline} because our chosen ADaPT baseline is enough strong and substantially outperforms them (see Table~\ref{tab:more_prompt} for reference).
\begin{table}[tbp]
\setlength\tabcolsep{3pt}
\centering
\resizebox{\linewidth}{!}{
  \begin{tabular}{lcccc}
  \toprule
    \multirow{2}{*}{\textbf{Method}}&\multicolumn{2}{c}{\textbf{ALFWorld}} &\multirow{2}{*}{\textbf{WebShop}}&\multirow{2}{*}{\textbf{TextCraft}}\\\
    &Seen&UnSeen&&\\
    \midrule
    Plan\&Solve\small{+GPT-3.5}&46.4&43.3&61.8&22.0\\
    Reflexison\small{+GPT-3.5}&56.4&57.5&62.4&25.0\\
    ADaPT\small{+GPT-3.5}&\underline{70.3}&\underline{71.6}&\underline{62.7}&\underline{26.0}\\
    \midrule
    \textbf{PGPO}\small{+Llama-2-7B}&\textbf{76.4}&\textbf{76.9}&\textbf{72.2}&\textbf{43.0}\\
  \bottomrule
  \end{tabular}
}
  \caption{\label{tab:more_prompt}Comparative experiments on PGPO with more prompt-based baselines. The best and second-best results are marked in \textbf{bold} and \underline{underline}.}
\end{table}

\subsection{For Benchmark Evaluation}\label{app:eval}
Our evaluation framework follows previous works~\cite{song-etal-2024-trial,xiong2024ipr} and extends it to TextCraft benchmark. 
The maximum number of steps for ALFWorld, WebShop and TextCraft is set to 20, 10 and 20, respectively.

\section{Additional Experimental Results}

\begin{table}[tbp]
    \centering
    \resizebox{0.9\linewidth}{!}{
    \begin{tabular}{cccc}
   \toprule
    \multirow{2}{*}{\textbf{Method}}&\multicolumn{2}{c}{\textbf{ALFWorld}} &\multirow{2}{*}{\textbf{WebShop}}\\\
    &Seen&UnSeen&\\
    \midrule
     \textit{w/o Plan}  & 72.1&68.7&61.8\\
     \textit{w/ NL Plan}&70.7&69.4&63.0\\
     \textit{w/ P-code Plan}&\textbf{75.0}&\textbf{72.4}&\textbf{63.6}\\
    \bottomrule
    \end{tabular}}
    \caption{Comparative experiments using P-code Plans generated by the model itself.}
    \label{tab:self-generated}
\end{table}

\subsection{Using Self-generated P-code Plan}
As described in Section~\ref{sec:plan_gen}, we use one powerful closed-source LLM (i.e., GPT-4o) to generate P-code Plans from existing ReAct-style datasets for good quality control. 
To further demonstrate the format advantage of P-code Plan in agent reasoning, not knowledge distillation from other strong models, we conduct supplementary experiments that uses the plans generated by the model itself to construct new SFT data. 
Taking Mistral-7B as an example, Table~\ref{tab:self-generated} shows SFT \textit{w/ P-code Plan} maintains its advantage over \textit{w/o Plan} and \textit{w/ NL Plan}, even when utilizing self-generated plans. 
This proves that the enhanced agent reasoning capabilities  should be attributed to the structured nature of P-code Plan, rather than knowledge distillation from other models.

\subsection{Compared with Executable Code Format}\label{app:execode}
Similar to the generation of P-code Plan, we first meticulously curate few Execode Plan (standing for plans in executable code format) demonstrations and then utilize GPT-4o to synthesize the plan data via few-shot prompting. 
Taking ALFWorld as an example, we report the average reward of SFT \textit{w/ P-code Plan}, \textit{w/ Execode Plan} and \textit{w/o Plan} across four open-source LLMs in Table~\ref{tab:cmp_execode}. 
The results show that SFT \textit{w/ P-code Plan} maintains better performance than SFT \textit{w/ Execode Plan}.
Sometimes SFT \textit{w/ Execode Plan} even falls behind  SFT \textit{w/ Plan}.
Through error analysis, we attribute this to two reasons: 1) executable code generation is more challenging than natural language or pseudocode generation; 2) \textit{Execode Plan} is more verbose than \textit{P-code Plan}, which may introduce some noise information into subsequent reasoning. 
\begin{table}[tbp]
\setlength\tabcolsep{3pt}
  \centering
  \resizebox{\linewidth}{!}{
  \begin{tabular}{lcccc}
   \toprule
   \textbf{Setting}&\textbf{\makecell[c]{Llama-2\\-7B}}&\textbf{\makecell[c]{Llama-2\\-13B}}&\textbf{\makecell[c]{Llama-3\\-8B}}&\textbf{\makecell[c]{Mistral\\-7B}}\\
\midrule
\rowcolor{myblue0!40}\multicolumn{5}{l}{\textbf{ALFWorld-Seen}}\\
   \textit{w/o Plan}&60.0&67.1&67.1&72.1\\
   \textit{w/ Execode Plan}&62.1&60.0&67.1&61.4\\
   \textit{w/ P-code Plan}&\textbf{65.0}&\textbf{72.9}&\textbf{68.6}&\textbf{75.0}\\
\midrule
\rowcolor{myblue0!40}\multicolumn{5}{l}{\textbf{ALFWorld-Unseen}}\\
   \textit{w/o Plan}&67.2&67.9&72.4&68.7\\
   \textit{w/ Execode Plan}&68.7&63.4&64.9&64.2\\
   \textit{w/ P-code Plan}&\textbf{70.1}&\textbf{70.9}&\textbf{75.4}&\textbf{72.4}\\
   \bottomrule
  \end{tabular}
  }
  \caption{Comparison between w/ P-code Plan and w/o Plan, w/ Execode Plan on ALFWorld. \textbf{Bold} indicates the best results of each model.}
  \label{tab:cmp_execode}
\end{table}
\begin{table}[tbp]
    \centering
    \resizebox{0.8\linewidth}{!}{
    \begin{tabular}{ccc}
    \toprule
    \textbf{Method}&\textbf{Qwen2.5-7B}&\textbf{Qwen2.5-14B}\\
    \midrule
    SFT & 62.6&64.8\\
    PGPO$_{SFT}$ &65.3&65.9\\
    ETO &67.7&68.8\\
    PGPO &\textbf{70.7}&\textbf{72.3}\\
    \bottomrule
    \end{tabular}}
    \caption{Average reward on WebShop.}
    \label{tab:qwen}
\end{table}

\subsection{Results on Qwen Series}\label{app:qwen}
To further validate the generalization of our method on more models, we test PGPO on Qwen2.5-7B and Qwen2.5-14B model~\citep{qwen2.5}. 
Taking WebShop as an example, Table~\ref{tab:qwen} shows that when applying to Qwen2.5 series, our method still achieves the best performance.
PGPO$_{SFT}$ represents the SFT model in our pipeline, which still has an advantage over the naive SFT baseline due to the incorporation of our designed P-code Plans. 
Since IPR baseline requires enormous inference costs for step-wise reward estimation, we are unable to reproduce them due to limited time constraints. 
We leave comprehensive comparisons as future work.


\begin{table}[tbp]
\centering
\resizebox{0.78\linewidth}{!}{
  \begin{tabular}{ccccc}
  \toprule
    \multirow{3}{*}{\textbf{Method}}&\multicolumn{4}{c}{\textbf{ScienceWorld}}\\
    &\multicolumn{2}{c}{\textbf{Seen}} & \multicolumn{2}{c}{\textbf{Unseen}}\\
    &AR&SR(\%)&AR&SR(\%)\\
    \midrule
    SFT&67.7&70.1&52.4&57.8\\
    ETO&69.0&70.7&56.8&66.8\\
    IPR&70.2&70.6&54.4&61.6\\
    PGPO&\textbf{75.5}&\textbf{75.8}&\textbf{66.2}&\textbf{76.8}\\
  \bottomrule
  \end{tabular}
}
  \caption{\label{tab:sciworld}Evaluation results of PGPO and baselines on ScienceWorld. Experiments are based on Llama-2-7B. The evaluation metrics are the average reward (AR) and success rate (SR).}
\end{table}

\subsection{Evaluation on ScienceWorld}\label{app:sciworld}
ScienceWorld~\cite{wang2022scienceworld} is one agent benchmark for testing scientific reasoning abilities, which provides dense rewards from 0 to 1. 
For training, we use the expert trajectories collected from~\citet{song-etal-2024-trial}, comprising 1483 instances. 
For evaluation, we includes the development set (ScienceWorld-seen) with 194 seen scenarios and the test set (ScienceWorld-unseen) consisting of 211 new unseen task scenarios. The ScienceWorld-seen set can assess in-distribution capability while the ScienceWorld-unseen set can measure out-of-distribution generalization of the agents. 
Since its evaluation pipeline runs relatively slowly, we only conduct comparative experiments on the Llama2-7B model.
Table~\ref{tab:sciworld} shows the average reward and success rate of PGPO and baselines on ScienceWorld. 
We can observe that PGPO maintains better performance than ETO and IPR, further indicating the advantage of P-code Plan guidance.
Moreover, IPR baseline even falls short compared to the ETO baseline on ScienceWorld-Unseen, which once again confirms that step-wise reward does not necessarily elicit better LLM agents.


\begin{table}[tbp]
    \centering
\setlength\tabcolsep{4pt}
\resizebox{1.0\linewidth}{!}{
    \begin{tabular}{cccccc}
    \toprule
    \textbf{Dataset Size}&	10\%&	30\%&	50\%&	70\%&	100\%\\
    \midrule
    \textbf{ALFWorld-Seen}&	55.0&	70.7&	71.4&	75.0&	76.4\\
    \textbf{ALFWorld-Unseen}&	45.5&	61.9&	63.4&	69.4&	76.9\\
    \bottomrule
    \end{tabular}
}
    \setlength{\belowcaptionskip}{-5pt}
    \caption{Average reward on ALFWorld when PGPO applied on smaller training data subsets.}
    \label{tab:subset}
\end{table}

\subsection{Using Smaller Training Data Subsets}
We supplement the PGPO experiments about training Llama2-7B on smaller subsets (10\%, 30\%, 50\%, 70\%, 100\%). 
Taking ALFWorld benchmark as an example, the evaluation results are shown in Table~\ref{tab:subset}. 
The results indicate with 30\% of training data, the in-distribution performance of our method is already strong. 
However, the out-of-distribution performance continues to improve as the dataset size increases, demonstrating its scalability. 
This aligns with expectations since OOD tasks are more challenging, but with the full dataset, our method achieves significantly better performance.

\subsection{Collected Contrastive Trajectory Dataset Distribution over Iterations}\label{app:win_loss}
Figure~\ref{fig:win_loss} illustrates a connection between the trajectories collected and the performance of the agent over optimization iterations. 
It indicates that for the agent, maintaining an increasingly number of winning trajectories leads to improved average rewards during the optimization process. 
Thus, we can utilize this finding to select our optimal optimization iterations.

\section{Case Study}

\begin{figure*}[tbp]
    \includegraphics[width=1\linewidth]{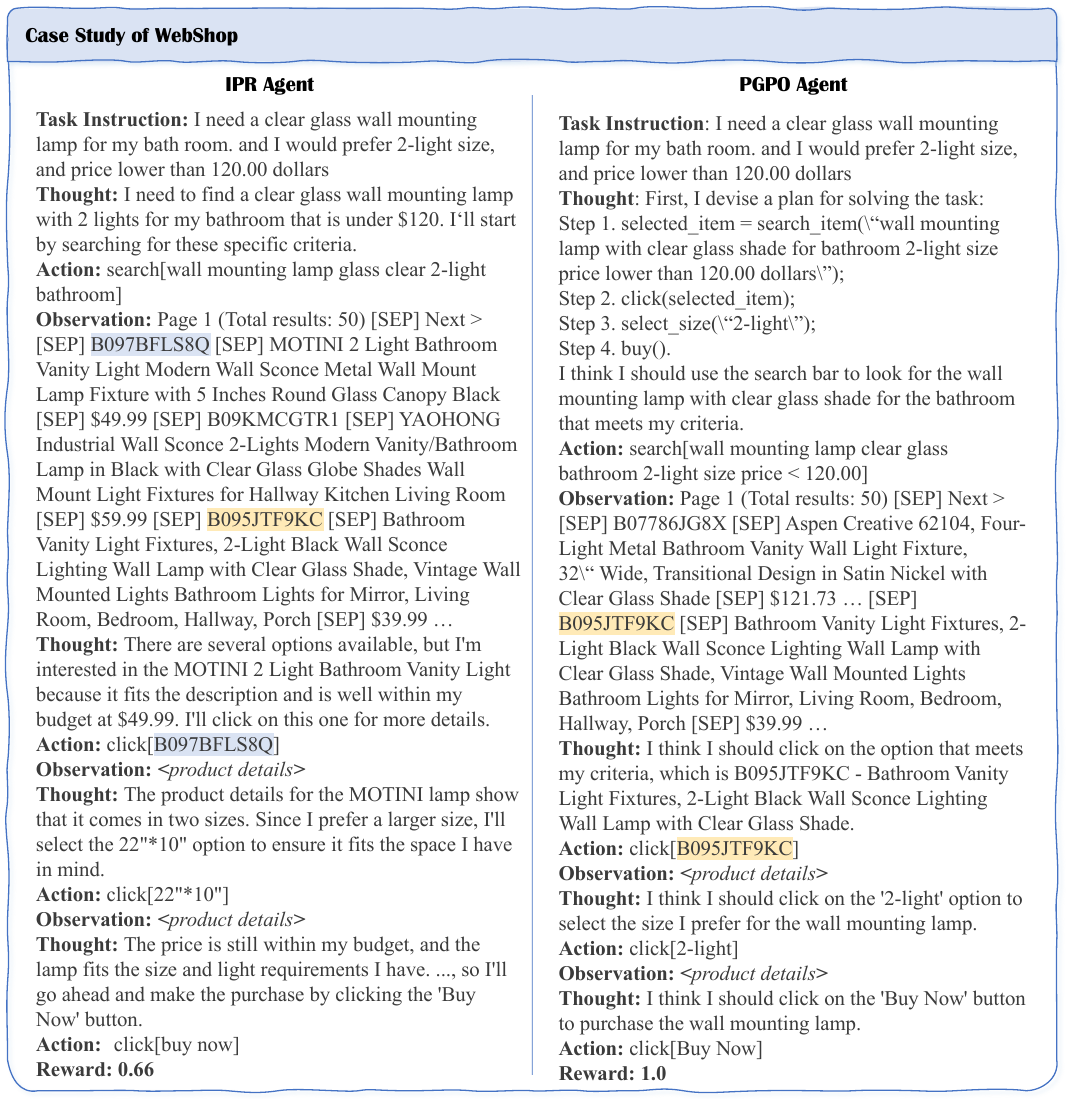}
    \caption{Case study for WebShop based on Llama-2-7B.}
    \label{fig:webshop_case}
\end{figure*}
\begin{figure*}[tbp]
    \includegraphics[width=1\linewidth]{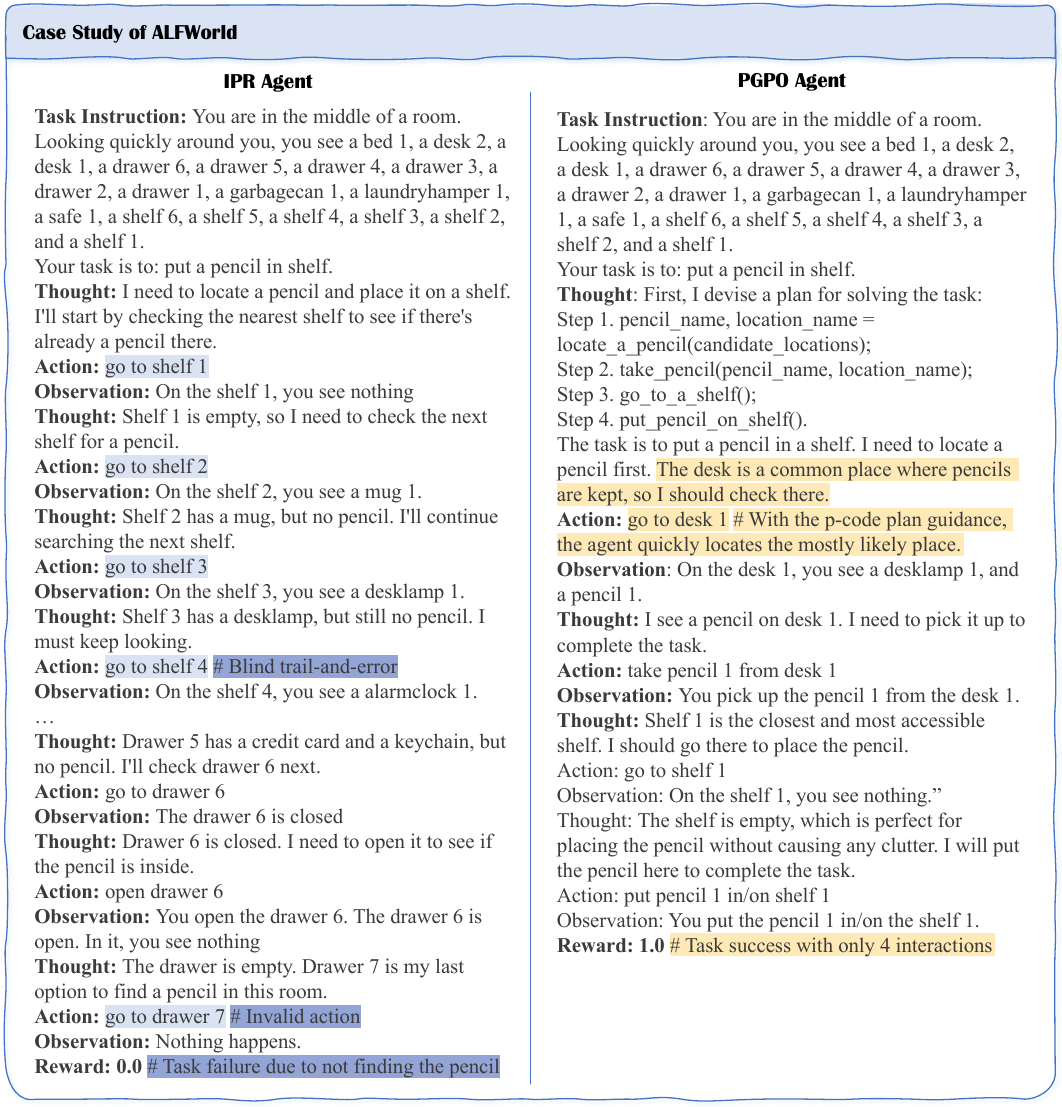}
    \caption{Case study for ALFWorld based on Llama-2-7B.}
    \label{fig:alfworld_case}
\end{figure*}
\begin{figure*}[tbp]
    \includegraphics[width=1\linewidth]{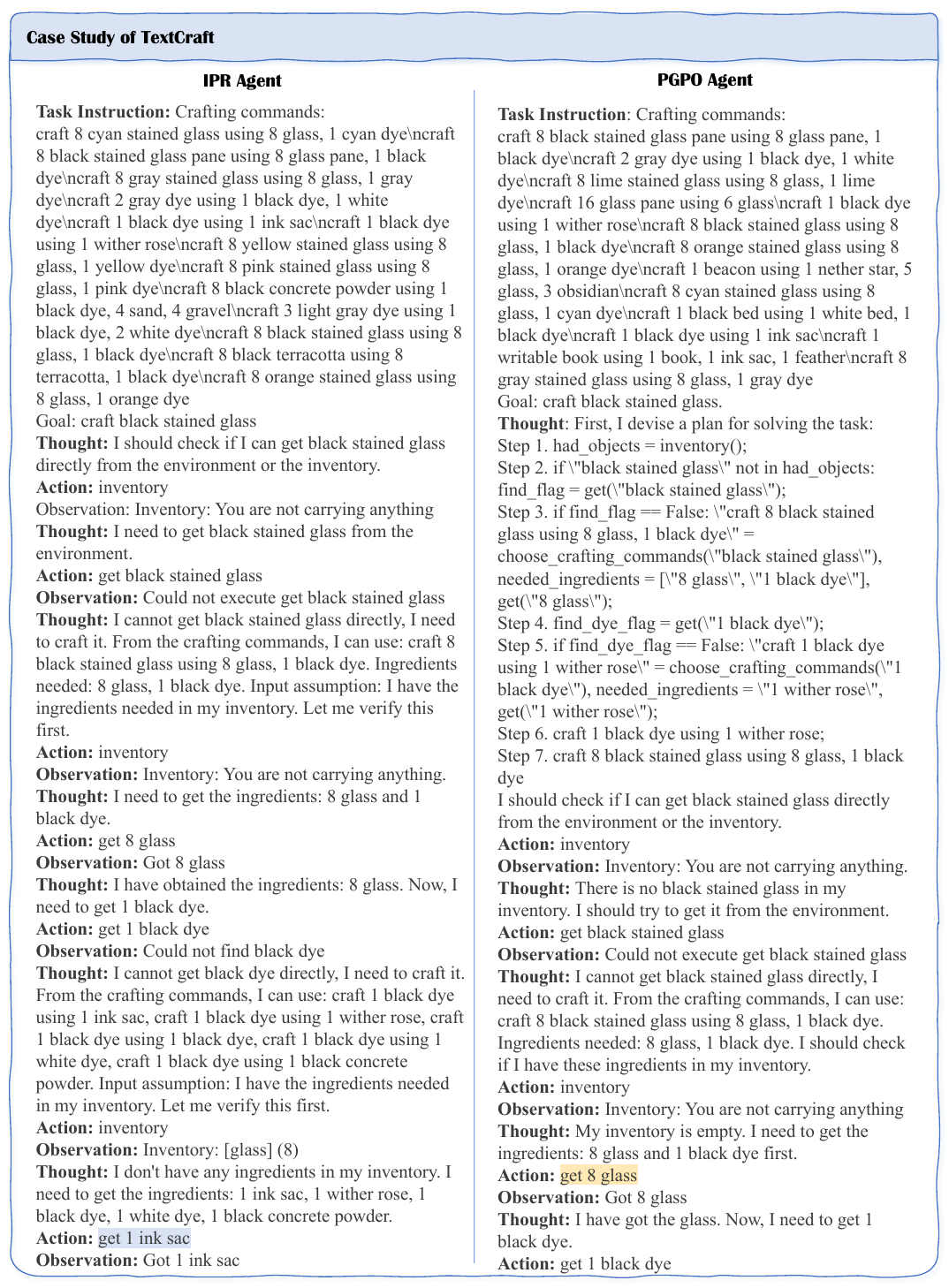}
    \caption{Case study for TextCraft based on Llama-2-7B.}
    \label{fig:textcraft_case}
\end{figure*}
\begin{figure*}[tbp]
    \includegraphics[width=1\linewidth]{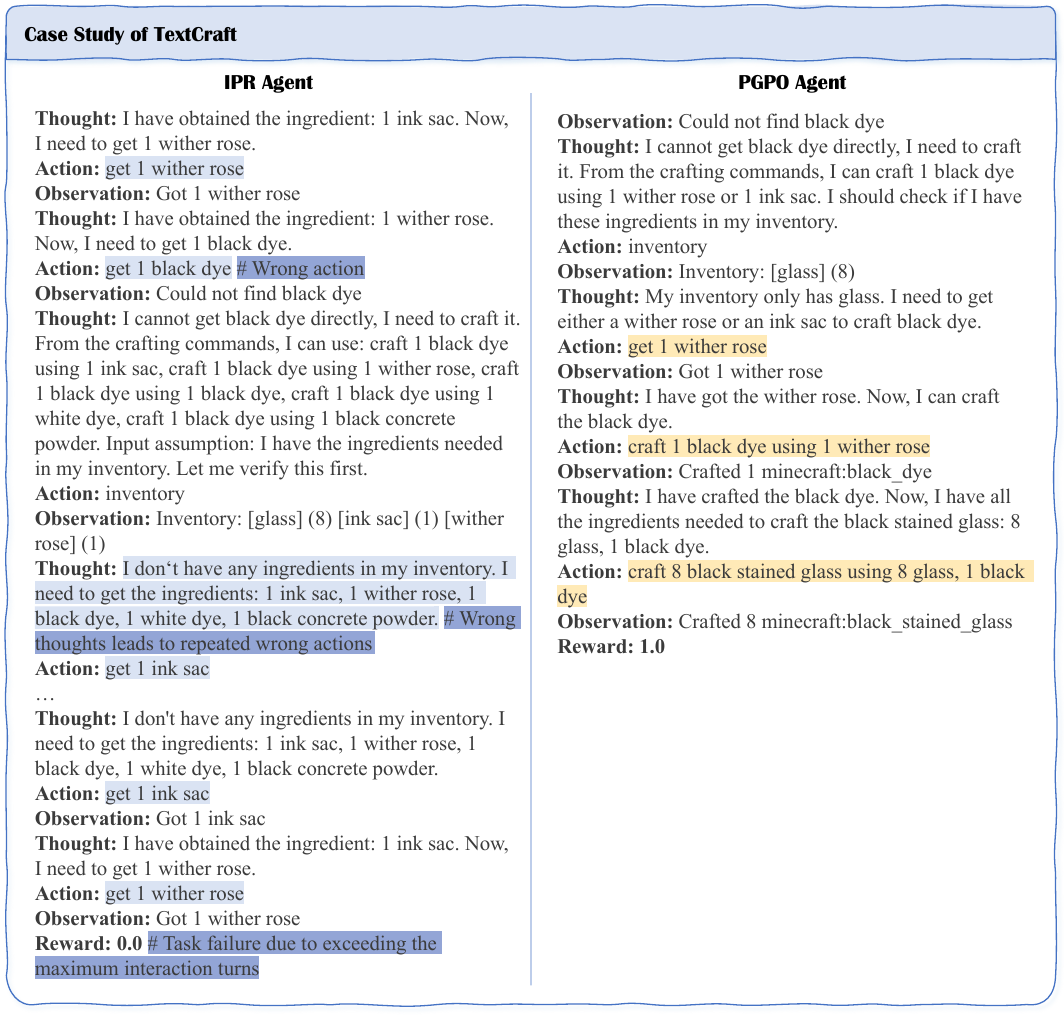}
    \caption{Case study for TextCraft based on Llama-2-7B (Continuations in Figure~\ref{fig:textcraft_case}).}
    \label{fig:textcraft_case_1}
\end{figure*}

We present one trajectory example for each agent benchmark compared with the current leading baseline IPR in Figure~\ref{fig:webshop_case},~\ref{fig:alfworld_case},~\ref{fig:textcraft_case} and~\ref{fig:textcraft_case_1}, respectively. 
\begin{itemize}[leftmargin=12pt,topsep=2pt,itemsep=-2pt]
    \item In the WebShop example, we find IPR baseline tends to select the product item located in the first position of one page, which is sub-optimal. 
    In contrast, our PGPO agent carefully browses through the whole page and successfully selects the optimal product located in the middle position of the page. 
    \item In the ALFWorld example, our PGPO agent quickly locates the pencil with the guidance of generated P-code Plan, thus completing the task with the minimum interaction turns. 
    Nevertheless, IPR agent blindly searches for the pencil in the shelves and drawers and fails to find the desired pencil due to exceeding the maximum interaction turns. 
    \item In the TextCraft example, we observe IPR agent made a mistake in the thinking process, thereby ending in a collapse.
    On the contrary, guided by the generated P-code Plan, PGPO agent efficiently complete the crafting task. 
\end{itemize}

\end{document}